\UseRawInputEncoding
\documentclass[final,3p,times]{elsarticle}

\usepackage{amssymb}

\usepackage{multirow}
\usepackage[table,xcdraw]{xcolor}
\usepackage{hyperref}
\usepackage{makecell}
\usepackage{graphicx}
\usepackage{caption}
\usepackage{lineno}
\usepackage{soul}
\usepackage{xcolor}
\usepackage{listings}
\usepackage{algpseudocode}
\usepackage[ruled,vlined]{algorithm2e}
\usepackage[autostyle, english = american]{csquotes}
\usepackage[english]{babel}
\MakeOuterQuote{"}

\journal{}
\makeatletter
\def\ifnopreprintline{\iftrue}
\makeatother

\begin{document}

\begin{frontmatter}



\title{Context-aware LLM-based AI Agents for Human-centered Energy Management Systems in Smart Buildings}


\author[inst1]{Tianzhi He}

\affiliation[inst1]{organization={School of Civil \& Environmental Engineering, and Construction Management, The University of Texas at San Antonio},
            addressline={BSE 1.310, One UTSA Circle}, 
            city={San Antonio},
            postcode={78249}, 
            state={TX},
            country={U.S.}}

\author[inst2]{Farrokh Jazizadeh}
\affiliation[inst2]{organization={Department of Civil and Environmental Engineering, Virginia Polytechnic Institute and State University},
            addressline={200 Patton Hall, 750 Drillfield}, 
            city={Blacksburg},
            postcode={24060}, 
            state={VA},
            country={U.S.}}

\begin{abstract}

This study presents a conceptual framework and a prototype assessment for Large Language Model (LLM)-based Building Energy Management System (BEMS) AI agents to facilitate context-aware energy management in smart buildings through natural language interaction. The proposed framework comprises three modules: perception (sensing), central control (brain), and action (actuation and user interaction), forming a closed feedback loop that captures, analyzes, and interprets energy data to respond intelligently to user queries and manage connected appliances. By leveraging the autonomous data analytics capabilities of LLMs, the BEMS AI agent seeks to offer context-aware insights into energy consumption, cost prediction, and device scheduling, thereby addressing limitations in existing energy management systems. The prototype's performance was evaluated using 120 user queries across four distinct real-world residential energy datasets and different evaluation metrics, including latency, functionality, capability, accuracy, and cost-effectiveness. The generalizability of the framework was demonstrated using ANOVA tests. The results revealed promising performance, measured by response accuracy in device control (86\%), memory-related tasks (97\%), scheduling and automation (74\%), and energy analysis (77\%), while more complex cost estimation tasks highlighted areas for improvement with an accuracy of 49\%. This benchmarking study moves toward formalizing the assessment of LLM-based BEMS AI agents and identifying future research directions, emphasizing the trade-off between response accuracy and computational efficiency.

\end{abstract}


\begin{keyword}
Large Language Model \sep Building Energy Management System \sep AI Agents \sep Human-Building Interaction
\end{keyword}

\end{frontmatter}

\section{Introduction}

Buildings account for approximately 30\% of global energy consumption \cite{manic2016building}, and are therefore critical in achieving energy efficiency. Targeted mitigation strategies, such as adaptive and human-centered control strategies \cite{jung2019human}, smart grid operations \cite{shi2020artificial}, Internet of Energy (IoE) adoption \cite{hannan2018review}, and fault detection frameworks \cite{gu2024digital}, combined with reactive and proactive user interactions, can facilitate efficient energy use in buildings \cite{hannan2018review}. To this end, Building Energy Management Systems (BEMS) have been developed and extensively studied \cite{hannan2018review}. These systems could optimize energy efficiency and support tasks such as human-centered operations, demand response, consumption and cost monitoring, and anomaly detection \cite{hannan2018review, mariano2021review}. Residential BEMS alternatives serve as intelligent control centers in IoT-enabled smart homes to coordinate meters, sensors, actuators, smart appliances, and hubs. BEMS' design concepts center around offering user-friendly interfaces for end users to manage energy use and achieve comfort and efficiency goals \cite{mahapatra2022home}. In different contexts, users can rely on different interaction modalities. Recent advances in AI-powered technologies, especially those involving natural language processing, have improved user-BEMS interfaces. A prominent example is the use of smart speakers with virtual assistants, which serve as intelligent hubs for both user engagement and smart appliance control \cite{he2022ai}.

Previous studies have broadly investigated the applications and architectures of BEMS with various objectives, such as enhancing energy efficiency, reducing operational cost, load profiling, and improving user comfort \cite{han2023home}. However, there have been challenges associated with their practical implementation in meeting such goals due to system complexities that may present a steep learning curve for users \cite{baedeker2020interactive}. These challenges can be observed in both commercial and residential settings, although they may be more pronounced in the latter due to users' varying levels of technical knowledge. Users may lack awareness of all system features, such as the availability and functionality of smart sensors and devices, or the optimal ways to use those systems, and might have limited ability to use them effectively, given the lack of intuitive interfaces. Moreover, users' goals may vary widely, from reducing energy cost to enhancing comfort or optimizing the integration of distributed energy resources \cite{mahapatra2022home}. However, users are often unaware of their energy use and generation patterns, and the relationship between those patterns and their behavior and actions, which limits the extent to which BEMS capabilities can be leveraged to improve energy efficiency. 

Intuitive user interfaces could be leveraged to partially address these complexities by facilitating user-technology interactions and guiding users in leveraging the system's capabilities toward achieving their objectives. However, given the varied user objectives, the design of such interfaces calls for flexibility. Conventional interfaces have commonly struggled to adapt to these varied objectives, often failing to accommodate open-ended and general user goals or support complex decision-making processes \cite{jin2023human}. In recent years, AI-powered technologies have emerged with the aim of facilitating user interactions through natural language interfaces. However, they often provide only basic support \cite{avdic2020intelligibility}. For instance, as noted, voice-enabled smart speakers have become prevalent but are commonly limited to answering simple queries posed by users \cite{avdic2020intelligibility}. The emergence of large language models (LLMs) has created opportunities to address those limitations. However, even when these interfaces are equipped with more advanced technologies, such as LLMs to understand and respond to user queries, their responses remain generic, whereas effective energy management, particularly when aligned with heterogeneous user objectives, requires awareness of building-specific hardware configurations, operational data patterns, and system attributes \cite{han2023home}. To address these limitations, our research investigates how LLM-based AI agents can be designed and configured to go beyond AI agents with strong conversation capabilities and function as ambient AI agents that are capable of providing context-aware support for users' goals. Context-awareness pertains to AI agent ability to understand the specifics of their operational environments, including energy consumption and generation patterns, and the operational states of appliances.

LLM-based agentic frameworks, exemplified by platforms such as OpenAI's ChatGPT series and Google's Gemini series, may offer potential solutions to the aforementioned challenges. Trained on large-scale textual datasets, LLMs learn language patterns and contextual relationships to understand, generate, and interact in human language \cite{brown2020language}. Researchers have successfully integrated LLMs into AI agents to enable autonomous reasoning steps (e.g., Chain-of-Thought prompting) that approximate certain aspects of human problem-solving and decision-making processes \cite{xi2023rise, wang2024survey}. With unified frameworks comprising key modules, such as profile (instructions for AI agents), memory, planning, and action \cite{wang2024survey} or perception, cognition (``brain''), and action \cite{xi2023rise}, different LLM-based AI agents have been designed to suit various tasks and applications. These include simulation agents in social science research \cite{kovavc2023socialai, li2023you}, tool-augmented agents in natural sciences \cite{boiko2023emergent, bran2023chemcrow}, and embodied agents for industrial automation and robotics \cite{mandi2023roco, rana2023sayplan, wu2023tidybot}. The integration of LLM-based ambient AI agents, capable of observing, perceiving, analyzing, and adjusting a BEMS' environment, has the potential to enable more flexible, convenient, and intuitive user interaction mechanisms between users and BEMS. Such mechanisms could, in turn, be leveraged to enhance user awareness and engagement in reactive and proactive building energy management systems and control applications \cite{king2024sasha}. Our vision centers around a BEMS AI Agent that can communicate with users in natural language, independently and autonomously analyze building data to identify context-aware responses to users' queries, and provide feedback and actions to help users achieve their energy management goals. To this end, our study (1) investigates a conceptual framework for LLM-based AI Agents, specifically tailored for context-aware smart building energy management, (2) develops a prototype of this framework for smart home energy management, and (3) evaluates its feasibility (in terms of effectiveness and latency), as well as generalizability in addressing several categories of user queries with varied complexities. This involved constructing a query benchmark and assessing how reliably the AI Agent can respond across different building conditions, providing a clear measure of its robustness and adaptability. We have sought to define the fundamental components, functionalities, and capabilities required for such a framework. For developing and evaluating our framework prototype, we utilized real-world smart residential building configurations and energy data from different households.

The rest of the paper is structured as follows. Section \ref{sec:LR} presents the research background and related studies. Section \ref{sec:framework} outlines the proposed framework, detailing the designed prototype, and the evaluation methods for the LLM-based BEMS AI Agent. The findings from our evaluations of the prototype on residential building energy datasets are detailed in Section \ref{sec:Findings}. This is followed by a discussion of the results and their implications for future work in Section \ref{sec:discussion}. Finally, Section \ref{sec:conclusion} concludes by presenting the findings and contributions of this study.

\section{Research Background}
\label{sec:LR}

\subsection{Building Energy Management Systems (BEMS) and Interfaces}

Building Energy Management Systems (BEMS) have evolved alongside advances in system design, information transmission technologies, and optimization algorithms over the past decade \cite{mariano2021review,han2023home}. In practice, a BEMS can be viewed as a modular system that brings together four major components: (1) sensors and meters that track environmental conditions and energy consumption, (2) communication networks and protocol translators (e.g., Zigbee), (3) data storage, forecasting tools, and optimization modules for environmental and energy analysis, and (4) control layers with user interfaces that connect to HVAC, lighting, energy information systems, and other controllable devices \cite{manic2016building,mahapatra2022home}. This layered architecture enables buildings to monitor and manage energy consumption, aligning efficiency with operational and comfort constraints \cite{mahapatra2022home,han2023home}. From a life cycle perspective, BEMS also supports operation and maintenance (O\&M) workflows in which anomaly detection, diagnostics, and performance tracking are increasingly data-driven \cite{chen2023review}. Effective energy management depends on rich, heterogeneous inputs that feed analysis and control modules, including historical energy use, the use of distributed energy resources, thermal dynamics, occupancy and schedules, operational constraints, and appliance controllability \cite{luo2019optimal,mariano2021review}. In residential contexts, BEMS commonly emphasize thermostatic control and HVAC scheduling, time of use shifting to reduce peak demand costs, and coordination of different resources such as rooftop photovoltaic (PV) systems and battery storage \cite{badar2022smart}. Increasingly, these approaches are being integrated with BEMS interfaces such as smart home energy hubs and appliance‐level control, allowing households to actively participate in energy management while maintaining comfort and minimizing operational costs \cite{blonsky2022home}.

BEMS interfaces have traditionally been designed around dashboard-oriented front ends that provide system alarms, time‐series trends, and setpoint controls to users. These visual dashboards are often optimized for expert users and typically offer limited interactivity or explanation beyond data display. In recent years, however, a shift has begun toward more occupant-centered interfaces that emphasize engagement, feedback, and actionable insights. Modular building energy systems demonstrate that embedding context-aware feedback, goal setting, and suggestions, can lead to energy efficiency while preserving comfort and productivity \cite{bonino2012home,jazizadeh2012human}. For instance, Francisco et al. integrated spatial, color‐coded Building Information Modeling (BIM) visualizations into feedback systems to help occupants better interpret energy use and meaningfully engage with building operations \cite{francisco2018occupant}. In residential contexts, Vassileva et al. evaluated how feedback devices (displays, dashboards) influence consumption when combined with contextual cues \cite{vassileva2013energy}. While effective for monitoring and supervisory control, such interfaces have rarely supported natural language-based interactions with occupants or context-aware guidance for specific end uses \cite{jin2017foresee}. Meanwhile, users increasingly face a growing number of smart devices without a unifying, intelligible control layer for energy management \cite{jin2017foresee}. Occupants could benefit from an intelligent central coordinating assistant to organize appliances and personalize interactions \cite{chen2017butler}.

Building on these observations, we summarize three major barriers for BEMS in residential settings. First, \textit{explainability}: most front-end systems have limited capacity to communicate the rationale behind control decisions, which reduces system explainability and undermines users’ ability to understand and potentially trust the BEMS. User studies consistently report dissatisfaction with building‐level feedback alone, highlighting a need for appliance‐specific insights and actionable recommendations \cite{nilsson2018smart}. Second, \textit{adaptivity}: current interfaces rarely tailor their content to users’ contexts, preferences or literacy levels. Energy literacy differs greatly across stakeholders, including operators, facility managers, and everyday occupants \cite{schwartz2015people}. Standardized kW/kWh displays often do little to engage or educate, whereas alternative framings such as monetary cost, comfort impact, historical comparisons, or social norms have been shown to be more effective \cite{buchanan2015question,nilsson2018smart}. Third, \textit{integration}: the heterogeneity of devices and fragmented ecosystems often prevents BEMS from providing unified control and delivering a coherent user experience. Case studies of residential energy management emphasize the importance of converting technical results into user-friendly recommendations and continuous operational guidance, rather than presenting them as one-time reports \cite{li2014insight, leitao2020survey}. Collectively, previous studies indicate that the success of a building energy management system is determined not only by the quality of its optimization algorithms but also by how users interact with the system. Building on this foundation, our study conceptualizes the BEMS not only as a control stack but also as a human‐centered interface that must coordinate algorithms, devices, and users, an orientation that directly motivates the AI agent design presented in this study.

\subsection{LLM-based AI Agents for Smart Buildings and BEMS}

Large Language Models (LLMs) are foundation models trained to understand and generate human language, supporting tasks such as question answering, summarization, translation, and content generation \cite{chang2024survey,radford2019language}. Beyond text, LLMs can integrate structured inputs and tool outputs, making them flexible for data retrieval, reasoning, and explanation \cite{liu2023summary}. Recent studies frame LLMs as the “brains” of interactive AI agents that couple language understanding with external tools and actuators \cite{xi2023rise,zhao2023depth}. Compared with rule-based agents, LLM-based AI agents can interpret vague intents, decompose tasks, call functions (e.g., data queries, simulations), and synthesize results into action plans or explanations \cite{zhao2023depth,jiang2024eplus}. They have been explored in scientific discovery \cite{li2023you,boiko2023emergent}, software engineering \cite{xi2023rise}, education \cite{liu2023summary}, and the AEC domain \cite{saka2023conversational}, often serving as a reasoning-and-orchestration layer that translates user goals into tool calls. In the context of energy and buildings, these capabilities suggest a path to smart building interfaces that interact with users in natural language while grounding responses in energy information and operational constraints \cite{jiang2024eplus}. Such interfaces can lower cognitive load, personalize feedback, and offer explanations of control choices \cite{michelon2025large}. However, practical deployment must also address accuracy, latency, and cost, and combine LLMs with guardrails and domain tools to ensure reliable, auditable operation in building environments \cite{liu2025large}. 

At the same time, the advancements of IoT devices have increased functionality but also complexity in smart environments \cite{guo2019review}. Users and operators frequently confront fragmented apps and rigid rules (e.g., If This Then That (IFTTT) conditional statements for automation) that struggle with nuanced intents and evolving contexts \cite{king2023get}. LLM-based AI agents address this by performing contextual reasoning over sensor states, schedules, and user preferences, enabling fluent, natural interactions (e.g., “It feels stuffy in the room, can you fix it?”) and generating concrete actions such as adjusting airflow and setpoints \cite{rivkin2024aiot}. As LLMs can generalize from language alone, they reduce the need for exhaustive command templates and enable faster implementation across heterogeneous device ecosystems \cite{king2024sasha}. Empirical studies lend strong support to these directions. For example, natural-language front ends have been shown to improve IoT platform usability when combined with semantic parsing and hierarchical planning \cite{rivkin2024aiot,xu2022building}. Similarly, datasets such as VISH expand the range of goal-oriented smart-home commands that agents can recognize and execute \cite{noura2020vish}. More recent systems go a step further by translating informal or vague statements into actionable multi-step plans, enabling coordination across multiple devices and constraints \cite{king2024sasha}.

LLM-based AI agents are also being adopted within BEMS, a domain historically reliant on manual or rule-based workflows and specialized interfaces \cite{zheng2023bim}. The AI agents align naturally into BEMS workflows, in which they can parse a user query, fetch historical data and constraints, evaluate scenarios, and present options and plans to users. By linking language understanding to data pipelines, agents can translate questions into analyses (e.g., “What drives last month’s peak?”), summarize anomalies, and recommend targeted actions such as thermostat setbacks, schedule tuning, or retrofit opportunities \cite{saka2023gpt}. Beyond conversational support, researchers report progress on LLM-enabled pipelines for load analysis, forecasting, and fault/anomaly detection \cite{jiang2024eplus,gu2025toward}, where statistical outputs are paired with narrative explanations that foster more effective interactions \cite{zhang2024automated}. These trends point toward a BEMS future in which optimization and interaction could co-exist within a single, conversational control interface. Several case studies illustrate the emerging applications. For example, Gamage et al. proposed a contextualized QA framework that uses ChatGPT to resolve ambiguous or incomplete user queries in energy management systems, streamlining occupant-system communication \cite{gamage2023augmenting}. Zhang et al. developed an automated, GPT-assisted data-mining workflow for building energy analysis and report 89.17\% accuracy in identifying chiller-plant waste patterns over a year-long dataset \cite{zhang2024automated}. Jin et al. translated natural-language intents into optimization problems to align recommendations with user-specific energy preferences (e.g., EV charging, rooftop PV) \cite{jin2023human}. Ahn et al. embedded ChatGPT within an EnergyPlus-in-the-loop setup to coordinate HVAC control for comfort and CO\textsubscript{2}, achieving simulated savings of 16.8\% \cite{ahn2023alternative}. Taken together, previous studies suggest that LLM-based agents can make building data more interpretable, recommendations more actionable, and controls more transparent. By anchoring the AI agents in the BEMS pipeline reviewed in prior work and explicitly targeting current BEMS interface shortcomings, the goal of this study is to advance a practical pathway for an LLM-enabled, human-centered AI agent for energy management that is both accurate and effective in deployments.

\section{Methodology}
\label{sec:framework}

The methodology section describes the rationale behind the design of the conceptual framework, as well as the processes involved in the development of the prototype and its evaluation experiments using real-world data. 

\subsection{Conceptual Framework}

Development and key components of LLM-based AI agents from previous studies form the fundamental structure of the proposed BEMS AI Agent. In prior research \cite{xi2023rise, wang2024survey, zhao2023depth}, LLM-based AI agents commonly consist of three major modules, each containing several key components. The central processor module, or "brain" of the AI agent, is the first and most vital module \cite{xi2023rise}. Beyond the core LLM with its pretrained knowledge for reasoning and planning, the "brain" includes other key components, such as the system instructions and memory of the AI agent \cite{wang2024survey, zhao2023depth}. System instructions of the LLM define the roles and tasks of AI agents, usually in the form of text-based prompts that instruct and guide the agent's behavior \cite{wang2024survey}. Memory, comprising both short-term and long-term memories, helps the agents understand context, accumulate experiences, and perform tasks more efficiently and accurately \cite{wang2024survey, zhao2023depth}. 

The second module is the perception module, which processes multiple modalities of data sources, such as textual, visual, audio, and other structured data \cite{xi2023rise}. Functioning as the AI agent's primary interface with the environment, this module helps the agent to extract relevant features from sensory inputs and support environment representation and informed decision-making \cite{xi2023rise}. The third module, the Action module, is responsible for both traditional LLM-based actions, such as textual output, as well as expanded actions through external tool integration \cite{xi2023rise, wang2024survey, zhao2023depth}. Pretrained LLMs have inherent limitations, such as a lack of up-to-date knowledge and the propensity to generate hallucinated or factually inaccurate outputs and commands \cite{zhao2023depth, roller2020recipes}. To address these limitations, specialized external tools have been developed to expand the AI agent's action space, utilizing supplementary resources and functions to accomplish intended outcomes \cite{xi2023rise, wang2024survey, zhao2023depth}. 

\begin{figure}[ht]
\centering
\includegraphics[width = 1.0 \linewidth]{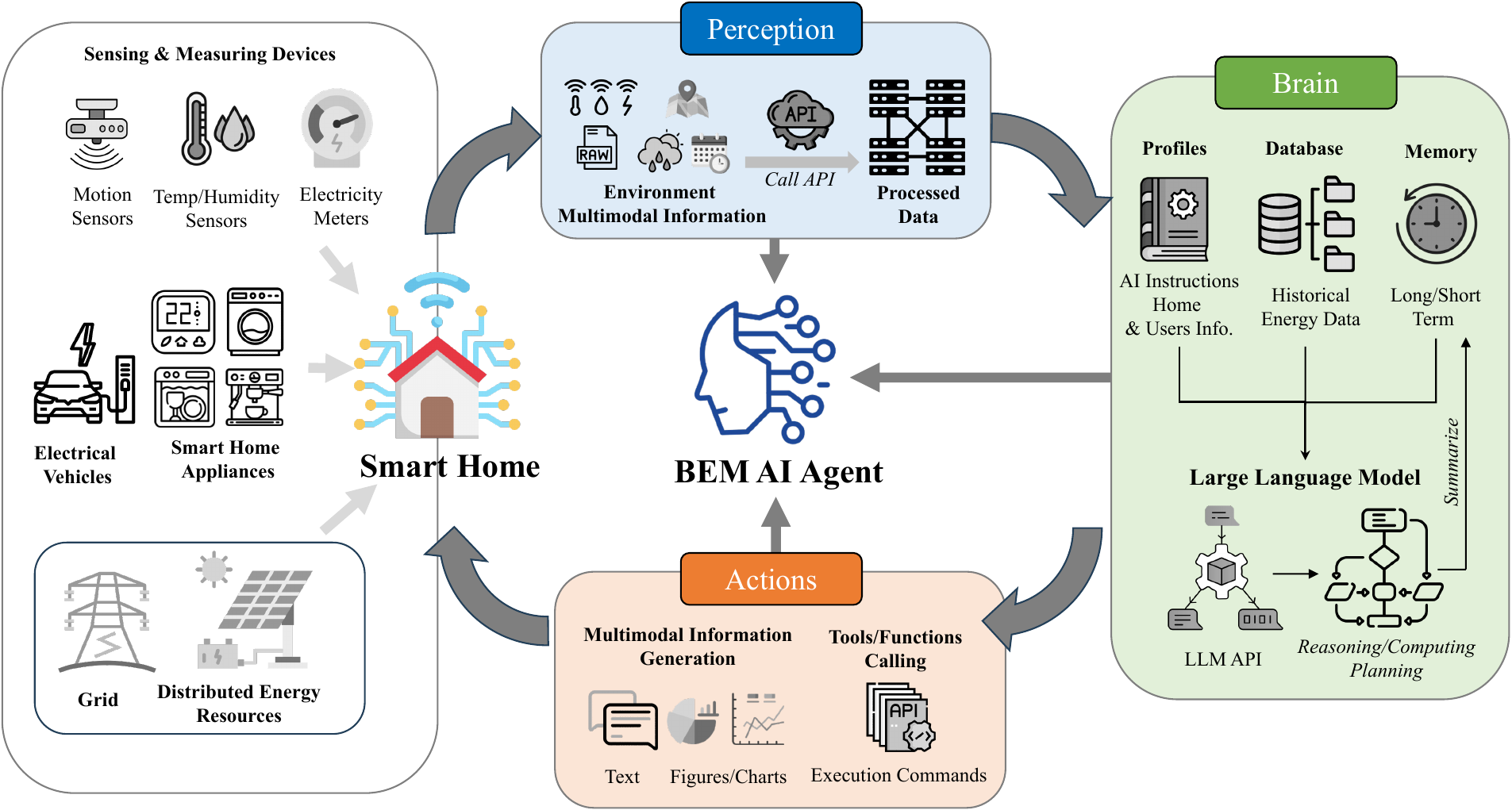}
\caption{The conceptual framework of the LLM-based Human-centered BEMS (Building Energy Management System) AI Agent}
\label{fig:framework}
\end{figure}

Drawing on these foundational modules and building on the established frameworks for Building Energy Management Systems (BEMS) \cite{han2023home, badar2022smart, chen2017butler, manic2016building, nilsson2018smart} and AI agents \cite{guo2019review, king2023get, xu2022building, noura2020vish, ahn2023alternative, zhang2024automated} from prior research, we have introduced a conceptual framework for the LLM-based AI agent for human-centered building energy management system, as illustrated in Figure \ref{fig:framework}. 

\subsubsection{Perception Module} 
The perception module forms the sensory foundation of the AI agent, enabling it to acquire and preprocess multimodal data from both indoor and outdoor environments. It interfaces various sensors and devices, such as smart electricity meters (to measure both aggregate and appliance level energy use \cite{jazizadeh2018embed,pecanstreet}), occupancy sensing systems, temperature and humidity sensors (either from a thermostat or from distributed nodes), indoor air quality (IAQ) sensors, wearable sensors, smart appliances, energy storage systems, and electric vehicles to collect contextual data critical for assessing energy usage, environmental conditions, and device states. Additional data streams, such as weather information, utility constraints, occupant location (indoor positioning of occupants), and calendar events, can also be gathered via specialized application programming interfaces (APIs) from third-party sensors and resources. These raw data are processed, structured, and relayed to the brain module to provide a comprehensive picture of the smart environment states either in real-time or as historical observations. Analogous to how sensory organs (e.g., eyes and ears) gather data from the environment and send it to the human brain for interpretation, the perception module of the ambient AI agent serves as the brain’s counterpart, integrating and interpreting data from various data streams to construct a coherent representation of the environment. It should be noted that not all environments are equipped with a full range of sensors and smart devices, and therefore, the ambient AI agent should be able to adapt and infer the state of the environment according to the available contextual configuration. Nonetheless, a standardized (ontology) representation of hardware configurations could facilitate uniform implementation of the perception module across different environments. However, the LLM capabilities could potentially be leveraged to facilitate the mapping between hardware configurations and data streams.

\subsubsection{Brain Module}

At the core of the AI agent is the brain module (i.e., central computational module), which functions as the primary computational layer responsible for coordinating the operations. It is responsible for multimodal data integration, reasoning, and interfacing with the external control layer for actuation. It leverages code generation and execution capabilities to synthesize contextual data and develop models, employs tools to activate memory retrieval, and manages computational processes to enable intelligent interactions and energy management. The core functionalities of LLM-based data analysis tools could enable agents to perform autonomous context-aware reasoning and continuous learning in dynamic environments.

\textbf{Data and Memory:} The processed multidimensional contextual data, such as the historical energy data, originating from the perception module, is stored and organized in the brain module's database. The brain module also stores and utilizes other information, such as the AI Agent system instructions and user interaction records. The information storage in the brain module can be conceptually compared to human memory. Therefore, the brain module’s memory is divided into long-term and short-term categories. Long-term memory encodes user-defined preferences for devices/appliances usage and schedules, historic energy consumption patterns, history of user interactions, and other disturbances such as occupancy trends (if available). Short-term memory includes data related to transient states such as the real-time devices/appliance statuses, and the ongoing user interactions/conversations context. Technologies such as vector databases and their associated tools (e.g., Milvus \cite{wang2021milvus} and FAISS \cite{douze2025faiss}) are commonly employed to support long-term semantic memory, enabling efficient similarity-based retrieval of historical user interactions, preferences, and contextual knowledge \cite{lewis2020retrieval}. In parallel, time-series databases (e.g., InfluxDB \cite{naqvi2017time}, TimescaleDB \cite{shah2022performance}) are widely used to store and query high-resolution historical building performance data, including energy consumption, device states, and environmental sensor readings. To provide a structured and machine-interpretable representation of building assets and their interrelationships, ontology-driven digital twinning approaches are also increasingly adopted. Standards and frameworks such as Brick Schema \cite{balaji2016brick} and BOT (Building Topology Ontology) \cite{rasmussen2020bot} enable semantic interoperability across heterogeneous building systems by formally modeling equipment, spaces, sensors, and their topological and semantic relationships. Overall, the brain module is not bound to a single implementation but can incorporate various data, and memory technologies, enabling extensibility and comparative evaluation in future work.

\textbf{AI Agent's Identity and Contextual Encoding:} To provide the AI agent with contextual understanding, the brain module maintains a structured profile or system instructions that define the AI agent’s operational constraints, functional objectives, and user-specific requirements. This profile, stored in text format (or other structured formats such as JSON), encodes: User-defined constraints and preferences (e.g., thermal comfort preferences), building metadata (e.g., HVAC zoning configurations, energy sources and storage integration (as applicable), appliance specifications), and operational policies governing the AI agent’s decision framework in energy management. By embedding these parameters, the AI agent ensures that generated recommendations and control actions align with user needs and building-specific attributes.

\textbf{Data-driven Reasoning and Control:} The LLM interfaces with the memory and profile representations to process them and perform reasoning, computation for inference and predictive analytics, and control/actuation signal generation. Using its advanced language model, the AI agent integrates data from various sources to generate and run specialized data analysis. The AI agent utilizes historical trends, real-time sensor streams, and environmental parameters to execute data analysis to find patterns, create summary statistics, predict patterns and detect anomalies and guide the responses related to energy management strategies. To this end, it dynamically generates executable scripts for data analysis and rule-based automation. Processed insights are translated into actionable commands for the Action Module, which executes device-level adjustments, automated scheduling, and adaptive user guidance.

\subsubsection{Action Module} 
The action module manages user interactions (both visual and conversational) and controls various smart devices and appliances to manage energy use and facilitate human-building interaction. It manages communication with users, generating multi-modal outputs such as text messages or notifications with energy suggestions, and visual representations of energy use patterns through various visualizations (such as figures, charts, and graphs) to help users better understand those patterns and identify areas for improvement. Beyond user engagement, the action module executes direct control over smart devices and appliances while managing distributed energy resources (DERs) through commands generated using the LLM’s tools and function-calling capabilities. This includes autonomous scheduling, dynamic load adjustments, and state-dependent configuration of devices such as HVAC systems, lighting, energy storage systems (batteries), and energy generation resources (e.g., photovoltaic arrays, if applicable). When enabled with additional functions, the system can also be used to optimize operations for improved energy efficiency and reduced operational costs.

In this conceptual framework, which consists of three fundamental modules, the AI agent and users interact within a feedback loop in an environment, creating an adaptive system that continually monitors and adjusts operations. Sensors collect environmental data, which the perception module processes and forwards to the brain module for analysis along with user input. The action module then communicates with users and implements decisions to adjust devices, with ongoing monitoring refining actions for energy efficiency and occupant comfort.

\subsection{AI Agent Prototype Design}

To assess the feasibility of the proposed LLM-based ambient AI Agent for energy management in buildings, we developed a prototype tailored for residential buildings. During development and testing, we focused primarily on the functionality and performance of the framework’s brain module, given that our primary objective was to examine the performance of LLM-based reasoning capabilities rather than the framework's interfaces. As such, to represent the interfaces between the brain and the perception modules, we utilized existing building energy data (from real-world field studies) and simulated smart building monitoring at the aggregate and appliance levels. Likewise, we examined how the control signals from the brain module interface with the action module through simulated adjustments in the states of devices and appliances. The following subsections outline the internal structure of the brain module and how it interfaces with the other modules. To develop this prototype, we have adopted the OpenAI Assistants API. 

\begin{figure}[ht]
\centering
\includegraphics[width = 1.0 \linewidth]{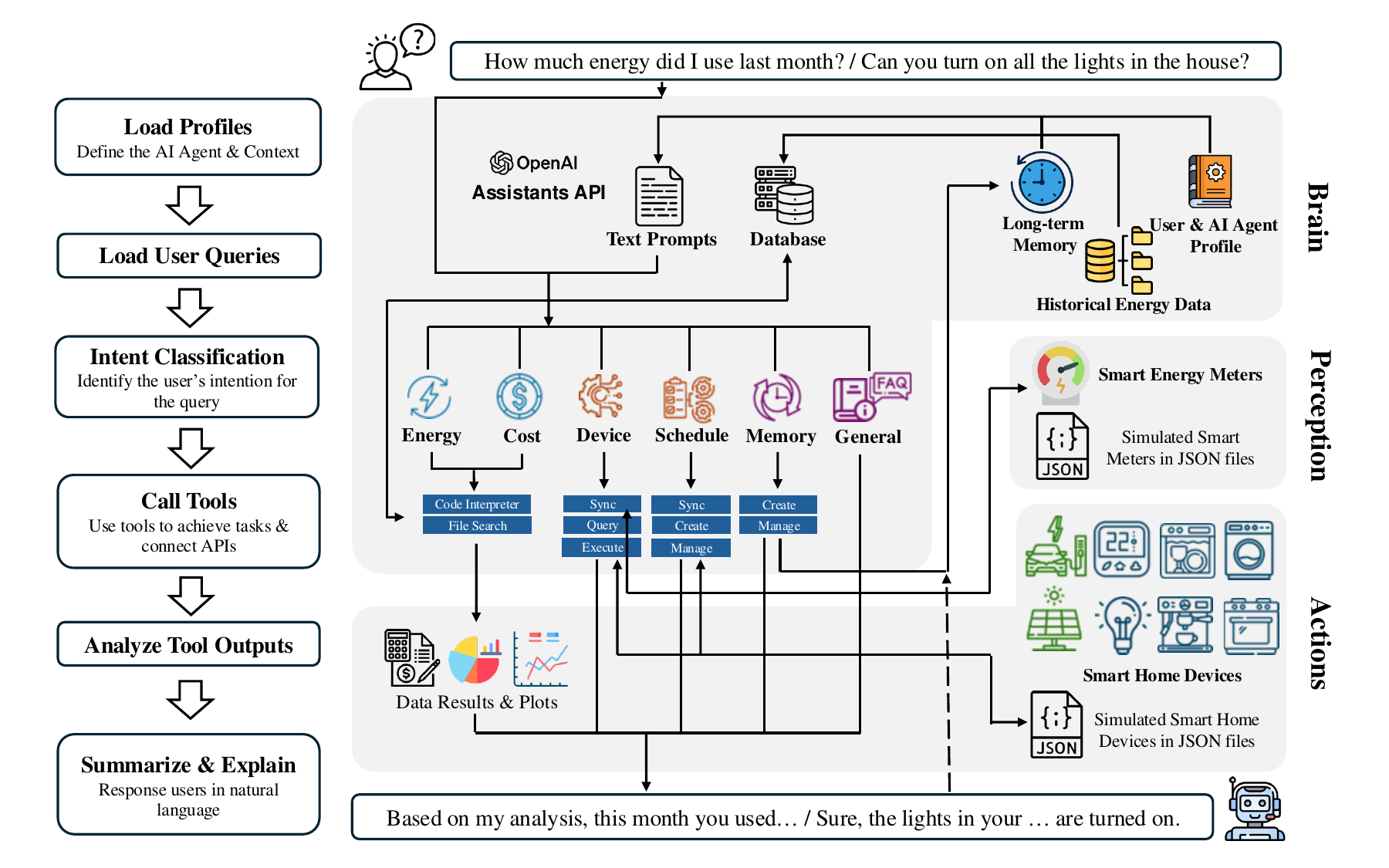}
\caption{The prototype design and operation flow of the LLM-based Building Energy Management System (BEMS) AI Agent 
}
\label{fig:flow}
\end{figure}

\subsubsection{Prototype's perception module} \label{subsec:prot. perc. module}

As noted, the perception module is the sensory infrastructure of the AI agent, which is responsible for collecting and organizing multimodal contextual data. Considering that our objectives include the assessment of the feasibility of autonomous computational analyses by the brain module and the evaluation of their generalizability to different contexts, for the perception module, we adopted the sensor system configuration as presented in the Pecan Street Project DataPort \cite{pecanstreet} and utilized their publicly available datasets as the primary source of contextual data. The Pecan Street Project is a smart grid initiative that deploys energy monitoring instrumentation (and partial actuation) in residential buildings across several U.S. states such as Texas, California, New York. Its infrastructure includes various sensing systems and energy management technologies---including energy management systems (EMS), rooftop photovoltaics (PV) meters, electric vehicle (EV) meters, and residential energy storage systems---to track and analyze electricity consumption and generation \cite{pecanstreet}. For measurement, the project uses eGauge energy sensors to monitor circuit-level electricity usage for multiple appliances and systems, providing one-minute circuit-level and whole-building electricity data from households \cite{pecanstreet, parson2015dataport}. In this study, we used historical data from four buildings in the Pecan Street DataPort to simulate the perception module’s functions. In the prototype, two separate files represent the simulated perception and brain modules interface:

\begin{itemize}
\item A JSON file (smart meters JSON file) stores the data on the current statuses of all devices and available resources. It contains date and time, building identifier (for different tested buildings), and available energy meters (i.e., available devices being monitored). Each meter entry includes an ID, name, description, status, unit (kW), and real-time energy consumption. For example: \{
  "Dishwasher": \{
    "name": "Dishwasher",
    "description": "eGauge meter data present for power draw of the dishwasher.",
    "status": "AVAILABLE",
    "online": true,
    "unit": "kW",
    "value": 1.8
  \}
\}.
This entry indicates that the dish washer is currently using 1.8 kW of power. When the AI agent needs current states of the devices/appliances (i.e., the “real-time” states), it calls functions to retrieve the relevant data from the smart meters JSON file, as illustrated in Figure \ref{fig:flow}.

\item Instead of a database system, in the developed prototype, a CSV file stores the historical energy consumption data for all devices/appliances, represented as energy use time series (in kWh). This file represents the processed historical energy data provided to the prototype's brain module. The CSV file headings include the names of different appliances and devices in the building in one table. The number of these devices is different across buildings.
\end{itemize}

\subsubsection{Brain module} \label{subsec:prot. brain module}

As the AI agent’s core computational processor that enables the overall system's functionality, the brain module encompasses several key elements including (i) profiles (for the AI agent, the user(s), and the building), (ii) the historical energy dataset, (iii) a long-term memory, (iv) a reasoning logic, and (v) several functions that enable analysis and actuation. The brain module elements and their integration are as illustrated in Figure \ref{fig:flow} and outlined in pseudocode in Algorithm \ref{alg:configure_bems}.

\textbf{AI Agent, User(s), and Building Profiles:} The profiles guide the AI Agent’s brain module by setting objectives and establishing its knowledge base. The \textit{AI agent} profile includes a general description of the AI agent’s role (e.g., "You are an AI agent for a smart building energy management system...") and its capabilities (e.g.,"You have access to the database of a smart building energy usage..."). This profile provides a comprehensive set of instructions for the AI agent and defines its identity and skill set. The \textit{user profiles} describe resident(s') characteristics (e.g., "The users who live in this single-family house are a family of four: 2 adults, 2 children...") These profiles can be expanded with more information to enable personalized user interactions. The \textit{building profile} comprises different sets of information: (1) the sensor list, which explains all sensors collecting energy consumption data from appliances, (2) the device list, which details all connected IoT smart appliances, and (3) the electricity time-of-use rates. The lists include names of the sensors and devices that match exactly with the data from the JSON files of the simulated appliances. The electricity rates include off-peak (00:00 - 17:00, 20:00 - 24:00) and peak (17:00 - 20:00) rates, credits for energy exported back to the grid, and EV discounts between 00:00 and 06:00. These information sets help the AI agent understand the context of the energy consumption data, specific device configurations, and associated energy cost to better assist users with their requests. All profiles are text-based prompts uploaded to the OpenAI Assistants API to build the AI agent.

\textbf{Energy Data:} The energy dataset serves as the input for the AI agent’s brain module, allowing it to analyze energy supply/consumption patterns and offer user suggestions. The energy data could include both historical and real-time data streams. As noted, for this prototype, we used data from the Pecan Street project in a CSV file format. To this end, we extracted one month of historical energy supply and consumption data from four houses (two in Texas and two in New York) as case study testbeds. These datasets consist of various time series sampled at 15-minute intervals, representing attributes such as grid electricity usage, energy generation, EV charging, and circuit-level appliance consumption data (e.g., air conditioning systems and kitchen appliances). Houses equipped with pv panel systems may exhibit negative grid usage values, indicating excess electricity export to the grid. 

\textbf{Long-term Memory:} The long-term memory is implemented as a persistent knowledge layer that captures stable user preferences, recurring behavioral patterns, and system-relevant constraints derived from user–agent interactions. Rather than storing raw conversational history, the memory formation process follows a deliberate extraction and consolidation mechanism, in which the AI agent identifies candidate memory entries based on explicit user instructions (e.g., “Remember that…”) or implicit signals indicating consistent and future-relevant intent. During this process, the AI agent transforms interaction content into structured, normalized memory representations that encode actionable information such as device targets, temporal conditions, and recurrence patterns. For example, when a user requests lowering the air conditioning (AC) temperature after 10 p.m., the agent derives a semantic rule such as “The user prefers a lower AC setpoint at 10:00 PM on a daily basis”, which is stored as a long-term memory entry. Stored memories subsequently inform future reasoning related to AC control, thermal comfort, and energy efficiency recommendations. Users retain direct governance over the memory system and may explicitly add, modify, or remove memory entries through natural-language commands. In the prototype, validated memories are persisted in a structured JSON format and accessed by the AI agent through tool-based retrieval and update operations. While this implementation is lightweight, it demonstrates a principled memory derivation mechanism that can be extended in future work using more advanced memory management frameworks.

\begin{algorithm}[htbp]
\caption{CONFIGURE\_BEMS\_AI\_AGENT}
\label{alg:configure_bems}
\begin{minipage}{0.92\linewidth}
\KwIn{
    AI\_agent\_profile, user\_profile, building\_profile (including energy\_meter\_profile, device\_profile, \\ energy\_rate); 
    historical\_energy\_data\_file (csv), memory\_file (json); 
    AI\_agent\_instructions, AI\_agent\_tools\_list, AI\_agent\_model}
\KwOut{
    Configured AI agent with integrated profile information, instructions, and tools}

\BlankLine
\textbf{Procedure:}

\begin{enumerate}
  \item \textbf{Initialize Agent Instructions}
    \begin{enumerate}
      \item Load AI\_agent\_profile, user\_profile, building\_profile (energy\_meter\_profile, device\_profile).
      \item Store these profiles in structured AI\_agent\_instructions.
    \end{enumerate}

  \item \textbf{Define Tools and Their Schemas}
    \begin{enumerate}
      \item Register available API tools (e.g., code\_interpreter, file\_search).
      \item Define additional tools (device, schedule, memory) with corresponding schemas and core methods (sync, query, execute).
    \end{enumerate}

  \item \textbf{Establish Reasoning Logic}
    \begin{enumerate}
      \item For each incoming user query:
        \begin{enumerate}
          \item Interpret query intention.
          \item Determine primary category (1 of 6 main categories).
          \item Identify secondary category (1 of 24 subcategories).
          \item Select functions from AI\_agent\_instructions:
          
            \Indp
                if query\_category = "Energy Consumption and Analysis - Energy Historical Data" then\;
                \Indp
                    code\_interpreter(historical\_energy\_data\_file)\;
                \Indm
                else if query\_category = "Device Control - Device General Operation" then\;
                \Indp
                    device\_execute(device\_status)\;
                \Indm

          $\vdots$ \Indm
        \end{enumerate}
      \item Check if additional information is needed:

        \Indp
                if additional information is required then\;
                \Indp
                    call memory functions or request user input\;
                \Indm
                if function call valid then\;
                \Indp
                    proceed\;
                \Indm
                else\;
                \Indp
                    respond with explanation\;
                \Indm  

    \end{enumerate}

  \item \textbf{Incorporate Agent Instructions}
    \begin{enumerate}
      \item Compile relevant profiles and reasoning logic in AI\_agent\_instructions.
      \item Specify AI\_agent\_model (e.g., "gpt-4o").
    \end{enumerate}

  \item \textbf{Prepare Data Files}
    \begin{enumerate}
      \item Upload historical\_energy\_data\_file.
      \item Include additional files (energy\_rate, memory\_file) as needed.
    \end{enumerate}

  \item \textbf{Deploy the AI Agent}
    \begin{enumerate}
      \item Deploy the configured BEMS\_AI\_agent via the appropriate API.
    \end{enumerate}
\end{enumerate}
\end{minipage}
\end{algorithm}

\textbf{Tools/Functions Design:} The AI agent has been equipped with various tools that enable it to perform complex tasks and interact with smart home sensors, meters, and devices. These tools facilitate the functioning of the perception and actions modules. Following intent classification, user queries are processed according to detailed instructions on the tools to be employed, corresponding to different query categories. OpenAI Assistants API provides two built-in tools that can be directly utilized: "File Search" for processing and searching through files, and the "Code Interpreter" for writing and executing Python code, enabling the processing of different data files. In addition to these built-in tools, we developed customized functions for interacting with smart home meters and devices. For meter-related tasks, we implemented "action-meters-QUERY" to retrieve readings and information from smart meters within the smart home. For device-related tasks, a sync-query-execute workflow was implemented to ensure that the AI agent operates with grounded device information (as shown in Figure \ref{fig:flow}), rather than hallucinating unavailable devices or unsupported functionalities. The "action-devices-SYNC" and "action-devices-QUERY" functions retrieve device metadata and real-time operational states from the smart home environment, while "action-devices-EXECUTE" issues validated commands to modify device states accordingly. Schedule-related functions include "action-schedule-CREATE", "action-schedule-SYNC", and "action-schedule-CHANGE" to create, synchronize, and modify device schedules. Similarly, memory-related functions, such as "action-memory-CREATE", "action-memory-SYNC", and "action-memory-CHANGE", manage the AI agent’s memory generation and management. All the customized functions have been used to simulate the information queries and device control by processing the JSON file representing smart devices and energy sensors.

\textbf{Reasoning Strategy:} LLM-based AI assistants could use different reasoning strategies to complete tasks. Among different alternatives, in our prototype, we have included Chain-of-Thought (CoT) and Tree-of-Thought (ToT) frameworks to guide the LLM’s reasoning process \cite{besta2024demystifying} because they provide complementary reasoning capabilities. Unlike the traditional Input-Output (IO) modality, where the LLM directly generates a final response upon receiving users’ queries, the CoT framework, introduced by Wei et al. \cite{wei2022chain}, uses intermediate reasoning steps, enhancing the LLM’s reasoning, computation, and planning capabilities in complex tasks such as building energy management \cite{besta2024demystifying}. Additionally, the ToT framework allows the LLM to branch at certain points in the reasoning process, enabling it to decompose user queries into subcategories, which are then addressed with specific instructions or tools \cite{long2023large}.

For the ToT-based user queries intent classification, we adopted a set of user intent categories synthesized from prior Human-AI interaction and agent design studies \cite{jin2023human, king2024sasha}, and further extended them to support energy-centric analysis tasks relevant to building energy management. As summarized in Table \ref{tab:intent_categories}, user queries are organized into six primary intent categories: Energy (Energy Consumption \& Analysis), Cost (Cost Management), Device (Device Status \& Control), Schedule (Device Scheduling \& Automation), Memory, and General (General Information \& Support). Each primary category is further decomposed into three to five secondary intent categories, resulting in a total of 24 intents. This hierarchical categorization is embedded within the CoT reasoning structure (Algorithm \ref{alg:configure_bems}, Procedure 3), where the initial reasoning step focuses on identifying the primary intent, followed by a secondary classification step that refines the user’s underlying goal. The introduction of secondary intent categories enables more precise intent disambiguation and improves the agent’s ability to map user queries to appropriate tools/functions or control actions. All intent definitions, category descriptions, and representative example queries are encoded as text-based prompt instructions and integrated into the AI agent profile.

As illustrated in Algorithm \ref{alg:configure_bems} (Procedure 3) and Figure \ref{fig:flow}, the AI agent processes user queries and generates responses using a step-wise Chain-of-Thought (CoT) reasoning framework, while the Tree-of-Thought (ToT) framework supports structured intent classification and action selection. Upon receiving a user query, the agent first analyzes the user’s intent and classifies it into one primary category (out of six) and one secondary category (out of 24), as defined in Table \ref{tab:intent_categories}. For example, when a user asks about the appliance with the highest energy use in the past month, the AI agent classifies the primary category as "Energy Consumption \& Analysis" and the secondary category as "Energy Historical Data". This categorization represents the reasoning branches within the ToT framework, which govern how the agent explores multiple solution pathways or requests further details. Once the intent categories are identified, the CoT instructions encoded in the agent profile guide the generation of intermediate reasoning steps (e.g., selecting the appropriate energy data source or determining the required analysis). The agent then invokes pre-defined functions or tools, such as a code interpreter for historical energy analysis or a device execution function for appliance control. Some tools generate outputs directly (e.g., numerical results or visualizations), while others rely on data retrieved from external APIs, such as smart meter readings or real-time device status. In the final CoT step, the agent synthesizes the tool outputs and presents the results to the user in natural language. This integrated reasoning workflow enables the agent to adapt effectively to user queries of varying complexity.

\subsubsection{Action module} \label{subsec:prot. action module}

\textbf{Simulated Smart Devices:} Similar to the approach used for the perception module, in this prototype, the action module was simulated by mimicking the operation of external APIs, represented by a JSON data schema (Figure \ref{fig:flow}). This setup includes smart devices, each represented through a 
\{"Parameter":"Value/State"\} format to represent the status of devices, encompassing "devices-info" and "devices-state". To enable control operations, Python functions were developed as a mock backend, reflecting how commands would be executed on physical devices. These functions process incoming requests from the AI agent, adjust the simulated states of smart devices, and update the JSON files accordingly.

\textbf{AI Agent Responses:} The AI agent’s outputs, ranging from suggestions for energy savings and device control strategies to analysis results and visualizations, are presented to users through an interactive and conversational interface. For instance, the agent may provide graphs illustrating historical energy consumption, provide textual summary information, or propose adjustments to improve usage. During testing, these responses were logged and employed for the framework evaluation, allowing us to assess how effectively the AI agent responds to user queries for managing smart building energy usage.

\subsection{Prototype Evaluation}
\label{sec:evaluation}

\subsubsection{Experiment Design}

The evaluations focused on two aspects: (1) the feasibility and latency of the BEMS AI agent prototype in responding to different categories of user queries and (2) to assess the generalizability of the framework across different contexts. Therefore, we used historical data from four residential buildings (houses) with varied configurations. As noted, historical energy data for these houses were extracted from the Pecan Street Dataset, focusing on one month of energy consumption data from two buildings in Texas (ID: TX-01 \& TX-02, January 2018) and two buildings in New York (ID: NY-01 \& NY-02, August 2019). Table \ref{tab:test-buildings} shows the characteristics of these houses, which represent different smart home configurations. All four cases included photovoltaic panels and electric vehicle charging, which make them prosumers and expand the scope of query assessments.

\captionsetup{font=normalsize}
\begin{table}[ht]
\centering
\caption{Prototype test building information}
\label{tab:test-buildings}
\resizebox{\textwidth}{!}{%
\begin{tabular}{ccccc}
\hline
\textbf{Test Building ID} &
  \textbf{TX-01} &
  \textbf{TX-02} &
  \textbf{NY-01} &
  \textbf{NY-02} \\ \hline
Building Type &
  Single-Family House &
  Single-Family House &
  Single-Family House &
  Single-Family House \\ \hline
Location &
  Austin, Texas &
  Austin, Texas &
  Brooktondale, New York &
  Ithaca, New York \\ \hline
Total Square Footage &
  1725 &
  2700 &
  1575 &
  1750 \\ \hline
Number of Occupants &
  3 Adults &
  \begin{tabular}[c]{@{}c@{}}2 Adults\\ 2 Children\end{tabular} &
  2 Adults &
  \begin{tabular}[c]{@{}c@{}}2 Adults\\  2 Children\end{tabular} \\ \hline
\begin{tabular}[c]{@{}c@{}}Number of Smart \\ Energy Meters\end{tabular} &
  18 &
  16 &
  12 &
  10 \\ \hline
\begin{tabular}[c]{@{}c@{}}Smart Appliances \\ Energy Meters\end{tabular} &
  \multicolumn{1}{l}{\begin{tabular}[c]{@{}l@{}}Electrical Grid \\ Photovoltaic System\\ Living Room\\ First Bathroom\\ Utility Room\\ Air Compressor\\ Electric Vehicle Charger\\ Clothes Washing Machine\\ Electricity Clothes Dryer\\ Natural Gas Clothes Dryer\\ Dishwasher\\ Furnace Air Handler\\ Kitchen App 1\\ Kitchen App 2 \\ Microwave\\ Oven\\ Refrigerator\\ Vent Hood\end{tabular}} &
  \multicolumn{1}{l}{\begin{tabular}[c]{@{}l@{}}Electrical Grid\\ Photovoltaic System\\ Air Compressor\\ Electric Vehicle Charger\\ Clothes Washing Machine\\ Dishwasher\\ Disposal\\ Electricity Clothes Dryer\\ Furnace Air Handler\\ Kitchen App 1\\ Kitchen App 2\\ Microwave\\ Oven  1\\ Oven  2\\ Refrigerator\\ Vent Hood\end{tabular}} &
  \multicolumn{1}{l}{\begin{tabular}[c]{@{}l@{}}Electrical Grid\\ Photovoltaic System\\ Air Compressor\\ Air Compressor 2\\ Waterheater\\ Electric Vehicle Charger\\ Cloth Washer and Dryer\\ Freezer\\ Kitchen App 1\\ Kitchen App 2\\ Well Pump\\ Garage\end{tabular}} &
  \multicolumn{1}{l}{\begin{tabular}[c]{@{}l@{}}Electrical Grid\\ Photovoltaic System\\ Heater\\ Waterheater\\ Electric Vehicle Charger\\ Well Pump\\ Range\\ Electricity Clothes Dryer\\ Kitchen App 1\\ Garage\end{tabular}} \\ \hline
\begin{tabular}[c]{@{}c@{}}Data Set\\ Time Frame\end{tabular} &
  \begin{tabular}[c]{@{}c@{}}January 2018 (31 Days)\\ Heating Season\end{tabular} &
  \begin{tabular}[c]{@{}c@{}}January 2018 (31 Days)\\ Heating Season\end{tabular} &
  \begin{tabular}[c]{@{}c@{}}June 2019 (30 Days)\\ Cooling Season\end{tabular} &
  \begin{tabular}[c]{@{}c@{}}June 2019 (30 Days)\\ Cooling Season\end{tabular} \\ \hline
Number of Data Points &
  2976 &
  2976 &
  2880 &
  2880 \\ \hline
\end{tabular}%
}
\end{table}

Based on previous studies \cite{jin2023human, king2024sasha}, user queries were categorized into six primary and twenty-four secondary categories, as detailed in Table \ref{tab:intent_categories}. For each secondary category, five distinct user queries were posed to the AI agent. This comprehensive testing approach allows for a multifaceted evaluation of the AI agent’s capabilities. Queries related to energy consumption analysis and cost management assess the AI agent’s core functionalities, including providing suggestions and information visualization to users for home energy management. Queries regarding device status and control test the AI agent’s perception and action modules, particularly its connectivity with external APIs in a simulated environment. Meanwhile, queries on device scheduling, automation, and memory evaluate the AI agent’s ability to create schedules and store memory based on general or vague user requests. Finally, general information and support queries assess the AI agent’s performance in providing guidance, troubleshooting, and technical support. Different types of queries initiate different pre-defined tools (functions) configured for the AI agent, and the tool/function call results are used by the AI agent for further actions as elaborated in Algorithm \ref{alg:agent_test}). In total, 120 queries across six primary categories were tested for each testbed building, resulting in 480 observations overall. The AI agent’s responses and their response latency (time), as well as token usage were recorded in markdown files (md format) as shown in the Algorithm \ref{alg:agent_test}. These interaction logs were used for performance assessment using predefined evaluation metrics.

\captionsetup{font=normalsize}
\begin{table}[ht]
\centering
\caption{Primary and secondary categories of intent classification with example queries}
\label{tab:intent_categories}
\renewcommand{\arraystretch}{1.2}  
\resizebox{\textwidth}{!}{%
\begin{tabular}{m{3cm} m{4cm} p{11cm}}  
\hline
\centering\textbf{Primary Category} & \centering\textbf{Secondary Category} & \textbf{Test User Query Examples} \\ \hline
\centering\multirow{5}{*}{\makecell{Energy \\ Consumption \\ \& Analysis}} & \centering Historical Energy Data & How much energy did I use last month? \\ \cline{2-3} 
\centering & \centering Energy Prediction & What is the predicted energy use for the next month? \\ \cline{2-3} 
\centering & \centering Energy Optimization & How can I reduce my energy consumption during peak hours? \\ \cline{2-3} 
\centering & \centering Energy Suggestions & Based on my past month energy use, can you give me some suggestions to save energy? \\ \cline{2-3} 
\centering & \centering Energy Visualization & Can you show me a pie chart of my energy energy use by device or system? \\ \hline
\centering\multirow{4}{*}{\makecell{Cost \\ Management}} & \centering Cost Information & How much did I spend on AC last month? \\ \cline{2-3} 
\centering & \centering Cost Prediction & How much money will I save from my PV panels next month? \\ \cline{2-3} 
\centering & \centering Cost Suggestions & Based on my past month energy cost, can you give me some suggestions to save money on energy? \\ \cline{2-3} 
\centering & \centering Cost Visualization & Show me the cost I spent on charging my car over the past month in a plot. \\ \hline
\centering\multirow{5}{*}{\makecell{Device Status \\ \& Control}} & \centering Meter Status Check & What is my PV panel meter reading? \\ \cline{2-3} 
\centering & \centering Device Status Check & Is the living room light currently on? \\ \cline{2-3} 
\centering & \centering Device General Operation & Set the AC to 20 degrees. \\ \cline{2-3} 
\centering & \centering Group Device Management & Turn off all kitchen appliances. \\ \cline{2-3} 
\centering & \centering Device Custom Configurations & Set the living room light to a brightness level good for reading. \\ \hline
\centering\multirow{4}{*}{\makecell{Device \\ Scheduling \\ \& Automation}} & \centering Schedule Information & Have I set a schedule for my car charger? \\ \cline{2-3} 
\centering & \centering General Scheduling & Turn on my coffee maker at 7 in the morning. \\ \cline{2-3} 
\centering & \centering Conditional Automation & If the dishwasher is on, keep the kitchen light on. \\ \cline{2-3} 
\centering & \centering Schedule Management & Remove the schedule I set for AC. \\ \hline
\centering\multirow{3}{*}{Memory} & \centering Memory Information & What device do I usually turn on at sunset time? \\ \cline{2-3} 
\centering & \centering Memory Creation & Remember that I usually like to have the fan on for my AC. \\ \cline{2-3} 
\centering & \centering Memory Management & Forget my preference for the AC fan mode settings. \\ \hline
\centering\multirow{3}{*}{\makecell{General \\ Information \\ \& Support}} & \centering System Guidance and Tutorials & Guide me through the process of controlling my smart devices. \\ \cline{2-3} 
\centering & \centering Troubleshooting and Technical Support & My kettle doesn't work, can you help me check it? \\ \cline{2-3} 
\centering & \centering FAQs and General Queries & What should I do if I want to add a new device to my smart network? \\ \hline
\end{tabular}%
}
\end{table}

\begin{algorithm}[t]
\caption{TEST\_BEMS\_AI\_AGENT}
\label{alg:agent_test}

\KwIn{
    queries\_dict (120 user queries across 6 categories); \\
    memory\_file (json); \\
    developed AI agent (AI\_agent\_name)
}
\KwOut{
    updated\_memory\_file (json); \\
    For each query: a log of the agent’s interaction 
    (function calls, execution time, token usage, response content); \\
    interaction\_log\_file (md) with final results for evaluation
}

\BlankLine
\textbf{Procedure:}

\begin{enumerate}
  \item \textbf{Initialize Memory}
    \begin{enumerate}
      \item Load memory\_file.
      \item Prepare additional\_instructions (e.g., extra memory content or configuration parameters).
    \end{enumerate}

  \item \textbf{Process User Queries}
    \begin{enumerate}
      \item For each query\_id in queries\_dict:
        \begin{enumerate}
          \item Retrieve user\_query = queries\_dict[query\_id].
          \item Call AI\_AGENT\_SERVER(AI\_agent\_name, user\_query, data\_file\_path, ...):
          
            \Indp
                (a) Start the timer. \\
                (b) Run the AI agent with user\_query. \\
                (c) While agent\_state is \texttt{"queued"} or \texttt{"in\_progress"}: \\
                \Indp 
                    wait.\\
                \Indm
                
                (d) If agent\_state = \texttt{"requires\_action"}: \\
                \Indp
                    perform function calls, gather function\_result, \\
                    and submit function\_result back to the AI agent. \\
                \Indm
                (e) Continue running the AI agent until agent\_state = \texttt{"end"}. \\
                (f) Stop the timer and record execution time and token usage. \\
                (g) Record function calls, execution time, token usage, and response content in query\_id\_interaction\_log\_file (md). \\
                (h) If necessary, update memory\_file and save as updated\_memory\_file \\
            \Indm
        \end{enumerate}
    \end{enumerate}

  \item \textbf{Finalize}
    \begin{enumerate}
      \item Print confirmation: \texttt{"All queries have been processed by the AI Agent."}
    \end{enumerate}
\end{enumerate}

\end{algorithm}

\textbf{Evaluation Metrics:}
The prototype’s performance was evaluated using metrics derived from previous studies \cite{chang2024survey, zhang2024generative, jin2023human}. The evaluation covered five key indicators, including latency, functionality, capability, accuracy, and cost, which will be described below. The detailed description and the dimensions of these metrics have been presented in Table \ref{tab:eval-anova}.

(1) Latency --- The response latency of the AI agent was measured in seconds, defined as the elapsed time between receiving a user query and generating a final natural-language response. This latency measurement captures the end-to-end processing time of the agent, including intent classification, intermediate reasoning, tool call, and response synthesis. Latency statistics were analyzed across different query categories to assess the agent’s responsiveness under varying task types and reasoning complexity. By comparing processing times among categories, this evaluation provides insight into how different reasoning paths and tool dependencies affect system responsiveness in practical usage scenarios.

(2) Functionality --- The AI agent’s intent classification performance was evaluated by comparing its predicted primary and secondary intent categories for each user query against predefined ground truth labels. Ground truth intent annotations were specified in advance for all test queries. Classification accuracy was used as the primary evaluation metric, providing insight into the agent’s ability to correctly interpret user intent, adhere to the prompt-based instructions, and follow the Tree-of-Thoughts reasoning structure when categorizing queries and selecting appropriate actions.

(3) Capability --- The AI agent’s ability to appropriately invoke external tools (e.g., "Code Interpreter" or "action-devices-EXECUTE") was evaluated during response generation. Tool activation accuracy was assessed through manual inspection of whether the agent correctly selected and executed the required tools for each query. Responses that activated all expected tools were classified as fully accurate (score = 1). Responses that activated incorrect tools or failed to activate any required tools were classified as incorrect (score = 0). Partial accuracy (score = 0.5) was assigned when only a subset of the expected tools was correctly activated.

(4) Accuracy --- To evaluate the reliability and accuracy of the AI agent, we constructed a benchmark dataset containing ground-truth response logic and correct outputs for each test query by manually processing the input data. The agent-generated responses were then compared against this benchmark to assess response accuracy. For queries involving code generation or numerical calculations, both the correctness and completeness of the generated code were examined. Additional evaluation criteria were applied to specific query types. For prediction tasks, the agent’s outputs were compared with ground-truth future energy consumption data. Visualization tasks were evaluated based on readability and informativeness. Response accuracy was also rated using a three-level scale: correct (1), partially correct (0.5), and incorrect (0). Responses were identified as correct when both the reasoning process and the final outputs fully matched the benchmark logic and ground-truth results. On the other hand, a response was labeled as partially correct when portions of the reasoning process or underlying logic were valid, but the final numerical results or conclusions did not exactly match the ground-truth answers. Responses were labeled as incorrect when the reasoning, methodology, or final results substantially deviated from the benchmark, or when the generated outputs failed to address the query requirements.

(5) Cost --- The cost refers to the AI agent's inference cost, considering that the prototype used OpenAI’s GPT-4o in zero-shot mode. The number of tokens used (input and output) and the token usage cost associated with tool usage (e.g., “Code Interpreter” and “File Search”) were tracked. A token refers to a piece of text, such as a word or part of a word, that serves as the basic unit of analysis for the LLM's processing and understanding of language. Token costs were calculated based on OpenAI pricing models. It is worth noting that the price of the tokens may vary based on the usage of the language model and the updates of the APIs.

\textbf{Analysis Methods:} Descriptive statistics, including the mean, median, and standard deviation, were computed for all evaluation metrics, along with frequency distributions and ratios to summarize overall performance patterns. These statistics were visualized using bar charts and box plots to facilitate comparative analysis. Analysis of Variance (ANOVA) was employed to assess the generalizability of the proposed prototype across different buildings by examining whether statistically significant differences existed in evaluation metrics under varying building contexts. In addition, ANOVA was used to evaluate the AI agent’s performance across different query categories by analyzing between-group and within-group variance, thereby identifying whether certain types of queries posed systematically greater challenges to the agent. These analyses enable a rigorous examination of performance consistency and robustness across heterogeneous operational conditions. Complementary qualitative content analysis was conducted to provide deeper insights into the AI agent’s tool usage behavior and response generation strategies that are not fully captured by quantitative metrics.

\section{Results and findings}
\label{sec:Findings}

In this section, we have first presented the generalizability assessment of the AI agent across different buildings to show that the framework generalizes well across buildings and then discussed the performance of the proposed AI agent framework by using the average of evaluation metrics across all user queries.

\subsection{Framework's Generalizability}

To validate the framework’s generalizability, we calculated the mean value of the evaluation metrics for each testbed and conducted a one-way ANOVA test. The null hypothesis of the ANOVA test was that the means of the evaluation metrics from the four residential buildings are equal. The results are presented in Table \ref{tab:eval-anova}. As shown, all but one of the one-way ANOVA test results have a p-value greater than 0.05, indicating no statistically significant difference between the mean responses from the tests conducted on the four testbeds. The only metric showing a significant difference is the intent classification execution rate (p-value = 0.04), with a Tukey’s HSD post-hoc test revealing that the mean values from TX-02 and NY-02 testbeds were significantly different, with a mean difference of 0.16. These results demonstrate that the AI agent's responses across the four buildings are consistent, thereby verifying the generalizability of the framework for energy management tasks across different buildings with performance consistency.

Since the prototype's performance was consistent across the four testbeds, we averaged the performance metrics, as shown in Table \ref{tab:eval-anova}. In terms of \textit{latency}, the AI agent took approximately 23 seconds on average to respond to user queries. Regarding \textit{functionality}, around 60\% of the responses involved step-wise intent classification as guided by the Tree-of-Thought structure embedded in the AI agent's profile prompts. The intent classification achieved high accuracy (0.91) in identifying the first category of user queries, but a lower accuracy (0.75) for the secondary category classification task. As for the prototype's \textit{capability} in tool usage, the tool call accuracy was high at 0.94, with various tools (e.g., code interpreter, file search, and other functions) triggered based on the type of user query. The average \textit{response accuracy} among all four test houses of 0.79 indicates the AI agent's promising performance in handling user queries. From a \textit{cost} perspective, considering the OpenAI pricing for the tested model (gpt-4o, \$2.50 per 1M input tokens, \$10.00 per 1M output tokens, accessed in October 2024), the average cost per query was \$0.0776 (with a total of 29,467 tokens). The evaluation metrics vary across different categories of user queries, which will be discussed further in the following sections.

\begin{table}[ht]
\centering
\renewcommand{\arraystretch}{1.1}
\caption{Comparison of evaluation metrics for the prototype's performance across different testbeds including one-way ANOVA test results}
\label{tab:eval-anova}
\resizebox{\textwidth}{!}{%
\begin{tabular}{ccccccccc}
\hline
\multirow{2}{*}{\textbf{\begin{tabular}[c]{@{}c@{}}Evaluation\\ Metrics Categories\end{tabular}}} & \multirow{2}{*}{\textbf{\begin{tabular}[c]{@{}c@{}}Evaluation \\ Metrics\end{tabular}}} & \multicolumn{4}{c}{\textbf{Specific Test Houses (Avg.)}} & \multirow{2}{*}{\textbf{\begin{tabular}[c]{@{}c@{}}All Test\\ Houses (Avg.)\end{tabular}}} & \multicolumn{2}{c}{\textbf{\begin{tabular}[c]{@{}c@{}}One-way ANOVA \\ Test Results\end{tabular}}} \\ \cline{3-6} \cline{8-9} 
 &  & \textbf{NY-01} & \textbf{NY-02} & \textbf{TX-01} & \textbf{TX-02} &  & \textbf{F-Value} & \textbf{P-Value} \\ \hline
Latency & Process Time (Sec.) & 28 & 20 & 22 & 23 & 23 & 0.88 & 0.45 \\ \hline
\multirow{3}{*}{Functionality} & \begin{tabular}[c]{@{}c@{}}Intent Classification \\ Execution Rate\end{tabular} & 0.54 & 0.68 & 0.63 & 0.53 & 0.60 & 2.85 & 0.04* \\ \cline{2-9} 
 & \begin{tabular}[c]{@{}c@{}}Primary Category \\ Classification Accuracy\end{tabular} & 0.92 & 0.91 & 0.91 & 0.90 & 0.91 & 0.04 & 0.99 \\ \cline{2-9} 
 & \begin{tabular}[c]{@{}c@{}}Secondary Category\\ Classification Accuracy\end{tabular} & 0.78 & 0.70 & 0.80 & 0.75 & 0.76 & 0.71 & 0.55 \\ \hline
\multirow{5}{*}{Capability} & Tool Call Count & 2.19 & 2.24 & 2.53 & 2.33 & 2.32 & 0.58 & 0.63 \\ \cline{2-9} 
 & Code Interpreter Count & 0.92 & 0.98 & 0.96 & 0.94 & 0.95 & 0.04 & 0.99 \\ \cline{2-9} 
 & File Search Count & 0.14 & 0.16 & 0.13 & 0.13 & 0.14 & 0.20 & 0.90 \\ \cline{2-9} 
 & Other Function Count & 1.13 & 1.11 & 1.18 & 1.18 & 1.15 & 0.05 & 0.99 \\ \cline{2-9} 
 & Tool Call Accuracy & 0.96 & 0.96 & 0.91 & 0.91 & 0.94 & 1.72 & 0.16 \\ \hline
Accuracy & Response Accuracy & 0.80 & 0.77 & 0.80 & 0.78 & 0.79 & 0.34 & 0.80 \\ \hline
\multirow{3}{*}{Cost} & Prompt Token Num. & 28454 & 27815 & 30108 & 29372 & 28937 & 0.27 & 0.85 \\ \cline{2-9} 
 & Completion Token Num. & 551 & 528 & 507 & 533 & 530 & 0.16 & 0.92 \\ \cline{2-9} 
 & Total Token Num. & 29006 & 28342 & 30614 & 29904 & 29467 & 0.25 & 0.86 \\ \hline
\end{tabular}%
}
\vspace{2mm} 
\parbox{\linewidth}{\small *Note: The p-value is lower than the alpha level of .05, and the null hypothesis is rejected.}
\end{table}

\subsection{Performance Differences Among Categories of User Queries}

The performance evaluation metrics for different primary categories of user queries are presented in Table \ref{tab:metrics_function_capa} (presenting functionality and capability metrics) and Table \ref{tab:metrics_time_cost_acc} (presenting latency, cost, and accuracy). From the functionality perspective (i.e., intent classification and category identification), the prototype performed well in energy consumption \& analysis and cost management tasks, achieving high execution rates and query intent classification accuracy (Table \ref{tab:metrics_function_capa}). However, for queries related to IoT device control (device status \& control, device scheduling \& automation), the intent classification execution rate dropped to 0.55-0.62. The rate further decreased for tasks related to memory (0.13) and general information \& support (0.45). This suggests that, for complex tasks requiring a reasoning process, the AI agent follows the designed instructions to perform intent classification. However, for more straightforward user queries, such as general information \& support, the AI agent tends to bypass intent classification and directly processes the queries.

Regarding the AI agent’s capability in tool calls, it performed well with tool call accuracy exceeding 0.90 across all query categories. Queries related to the computation and calculation of energy and cost required more tool calls associated with the code interpreter (2.28 and 2.71, respectively). In contrast, queries related to smart device control more frequently triggered other function tools related to the control of IoT devices (2.63 for device status \& control, and 1.71 for device scheduling \& automation). The file search tool was primarily triggered for cost management tasks (0.8), where the AI agent needed to access stored files to retrieve energy pricing information for calculations.

\begin{figure}
\centering
\includegraphics[width = 1.0 \linewidth]{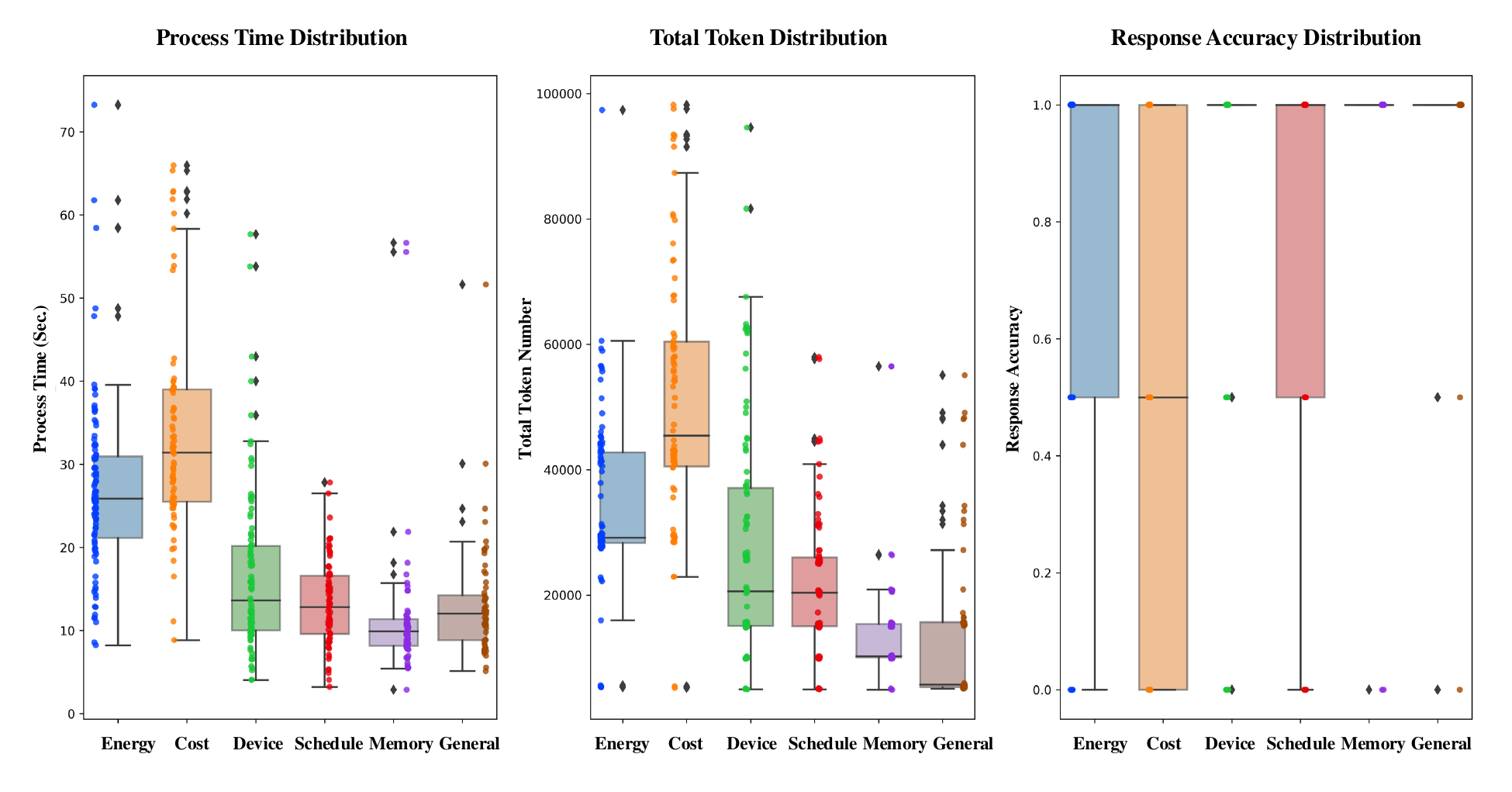}
\caption{Boxplots distribution of the evaluation metrics for different primary categories of the user queries}
\label{fig:boxplot}
\end{figure}

In terms of the prototype’s latency, queries related to cost management required the longest processing time, with an average of 49 seconds. Further investigation revealed two outlier values (over 600 seconds) that skewed the mean. After removing these outliers, the average processing time for cost management tasks was approximately 34 seconds. Energy consumption \& analysis tasks also required more processing time, with an average of 27 seconds. In comparison, the process time for device control and scheduling tasks were shorter, averaging 19 and 14 seconds, respectively, while memory and general information queries took 12 and 13 seconds. A major factor contributing to the longer processing times was the higher token usage in tasks requiring access to historical energy data and code execution for energy analysis and cost management. This is further illustrated in Figure \ref{fig:boxplot}, where the distribution of the evaluation metrics is presented in box plots for different categories of user queries. The box plots demonstrate a clear distribution pattern between processing time and token usage across query types. 

\begin{table}[ht]
\centering
\renewcommand{\arraystretch}{1.25}
\caption{Prototype's evaluation on functionality and capability on user queries in different categories}
\label{tab:metrics_function_capa}
\resizebox{\textwidth}{!}{%
\begin{tabular}{ccccccccc}
\hline
\textbf{\begin{tabular}[c]{@{}c@{}}Primary\\ Category\end{tabular}} & \textbf{\begin{tabular}[c]{@{}c@{}}Intent \\ Classification \\ Execution \\ Rate\end{tabular}} & \textbf{\begin{tabular}[c]{@{}c@{}}Primary \\ Category \\ Classification\\ Accuracy\end{tabular}} & \textbf{\begin{tabular}[c]{@{}c@{}}Secondary\\ Category\\ Classification\\ Accuracy\end{tabular}} & \textbf{\begin{tabular}[c]{@{}c@{}}Tool \\ Call\\ Count \end{tabular}} & \textbf{\begin{tabular}[c]{@{}c@{}}Code\\ Interpreter\\ Tool\\ Count \end{tabular}} & \textbf{\begin{tabular}[c]{@{}c@{}}File\\ Search\\ Tool\\ Count\end{tabular}} & \textbf{\begin{tabular}[c]{@{}c@{}}Other\\ Function\\ Tool\\ Count\end{tabular}} & \textbf{\begin{tabular}[c]{@{}c@{}}Tool\\ Call\\ Accuracy\end{tabular}} \\ \hline
Energy Consumption \& Analysis & 0.82 & 0.99 & 0.84 & 2.49 & 2.28 & 0.00 & 0.12 & 0.96 \\
Cost Management & 0.79 & 0.84 & 0.84 & 3.56 & 2.71 & 0.80 & 0.05 & 0.90 \\
Device Status \& Control & 0.62 & 0.95 & 0.63 & 2.86 & 0.08 & 0.00 & 2.63 & 0.90 \\
Device Scheduling \& Automation & 0.55 & 1.00 & 0.84 & 1.80 & 0.00 & 0.04 & 1.71 & 0.92 \\
Memory & 0.13 & 1.00 & 1.00 & 1.60 & 0.00 & 0.00 & 1.43 & 0.98 \\
General Information \& Support & 0.45 & 0.56 & 0.36 & 0.92 & 0.03 & 0.00 & 0.83 & 0.98 \\ \hline
\end{tabular}%
}
\end{table}

Although the AI agent used more resources and time for tasks related to energy and cost analysis, the response accuracy indicates that improvements are still needed. For simpler tasks such as device control and general information, the AI agent achieved high response accuracy (0.86 and 0.98, respectively). However, for more complex reasoning and analysis tasks, response accuracy dropped to 0.77 for energy consumption \& analysis queries and 0.49 for cost management queries. This suggests that while the BEMS AI agent can handle general energy management tasks, its reasoning capabilities, particularly for cost management, require further enhancement. The detailed response differences across query categories will be discussed in the following sections.

\begin{table}[ht]
\centering
\renewcommand{\arraystretch}{1.25}
\caption{Prototype's evaluation on latency, cost, and accuracy on user queries in different categories}
\label{tab:metrics_time_cost_acc}
\resizebox{0.85 \textwidth}{!}{%
\begin{tabular}{cccccc}
\hline
\textbf{\begin{tabular}[c]{@{}c@{}}Primary\\ Category\end{tabular}} & \textbf{\begin{tabular}[c]{@{}c@{}}Process\\ Time (Sec.)\end{tabular}} & \textbf{\begin{tabular}[c]{@{}c@{}}Prompt Token \\ Number\end{tabular}} & \textbf{\begin{tabular}[c]{@{}c@{}}Completion Token \\ Number\end{tabular}} & \textbf{\begin{tabular}[c]{@{}c@{}}Total Token \\ Number\end{tabular}} & \textbf{\begin{tabular}[c]{@{}c@{}}Response \\ Accuracy\end{tabular}} \\ \hline
Energy Consumption \& Analysis & 27 & 34771 & 854 & 35625 & 0.77 \\
Cost Management & 49 (34*) & 51523 & 1238 & 52761 & 0.49 \\
Device Status \& Control & 19 & 28956 & 231 & 29187 & 0.86 \\
Device Scheduling \& Automation & 14 & 21581 & 182 & 21763 & 0.74 \\
Memory & 12 & 14026 & 103 & 14130 & 0.97 \\
General Information \& Support & 13 & 13785 & 432 & 14217 & 0.98 \\ \hline
\end{tabular}%
}
\vspace{2mm} 
\parbox{\linewidth}{\small *Note: Two outliers (over 600 seconds of process time) were removed in the process time metric.}
\end{table}

\subsection{AI Agent Responses to User Queries}

\begin{figure}
\centering
\includegraphics[width = 1.0 \linewidth]{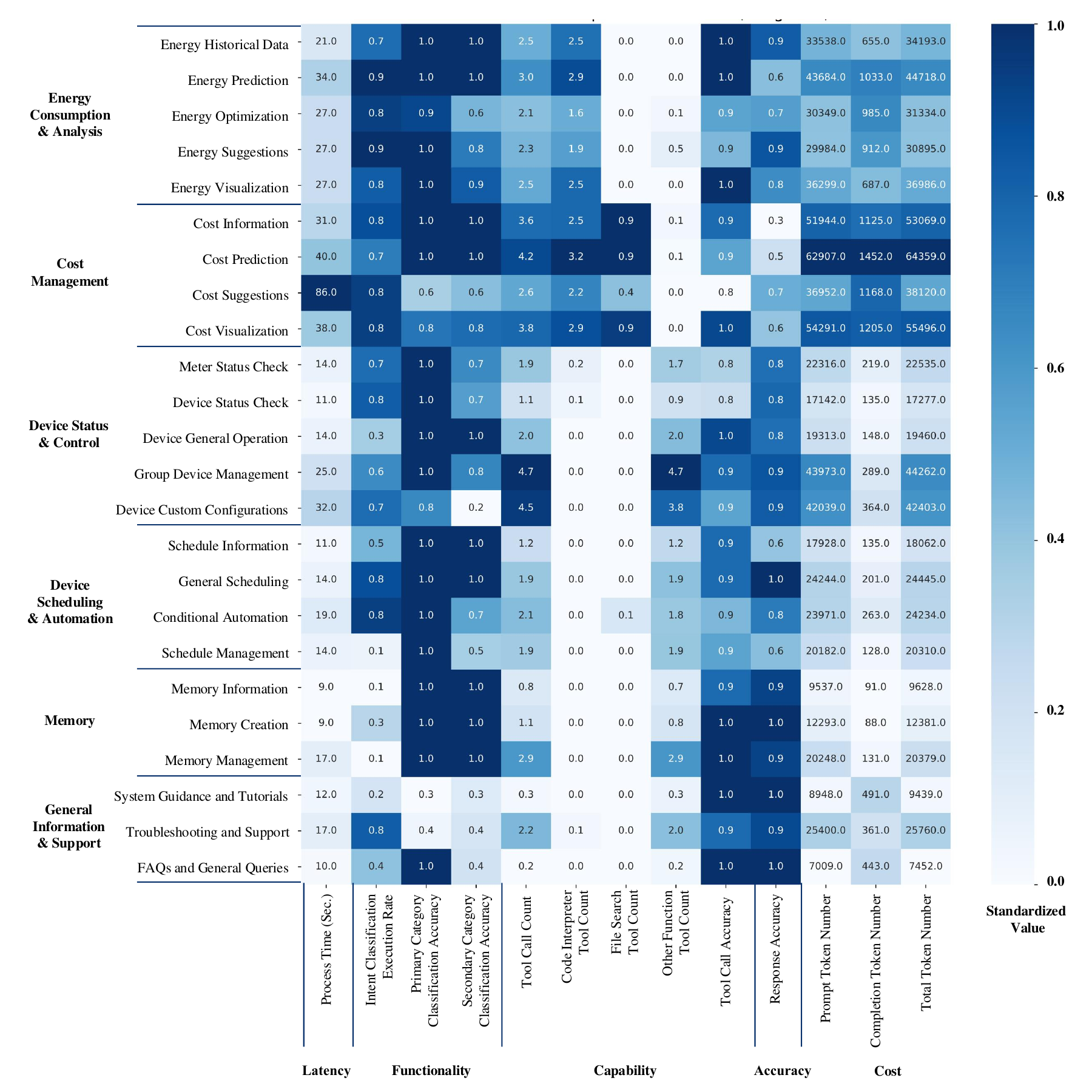}
\caption{Heatmap distribution of the evaluation metrics for different categories of tested user queries}
\label{fig:heatmap}
\end{figure}

To further evaluate the BEMS AI agent's performance in responding to different user queries, we compiled the mean values of the evaluation metrics across all secondary categories. Since each secondary category includes five queries, and these queries were tested across four residential buildings, each value represents the mean of 20 data points. These mean values were used to create a heatmap to visualize the distribution of the evaluation metrics for the various secondary categories of test queries, as shown in Figure \ref{fig:heatmap}. In the heatmap, the blue color indicates the relative values for each evaluation metric standardized along each column. The actual mean values of the evaluation metrics are displayed within each cell for reference. For example, in the "Latency - Process Time (Sec.)" column of Figure \ref{fig:heatmap}, within the "Device Status \& Control" primary category, tasks such as device and meter status checks take only around 11-14 seconds to execute. In contrast, more complex queries, such as those related to custom device configurations and group device management, take significantly longer, approximately 25-32 seconds. This example highlights the heatmap's usefulness in revealing the AI agent's performance variations when handling different types of user queries in greater detail. This visualization have also been referred to in subsequent subsections to explore the AI agent's dynamic response performance across various user queries.

\subsubsection{Energy Consumption \& Analysis}

\begin{figure}[t]
\centering
\includegraphics[width = 0.8 \linewidth]{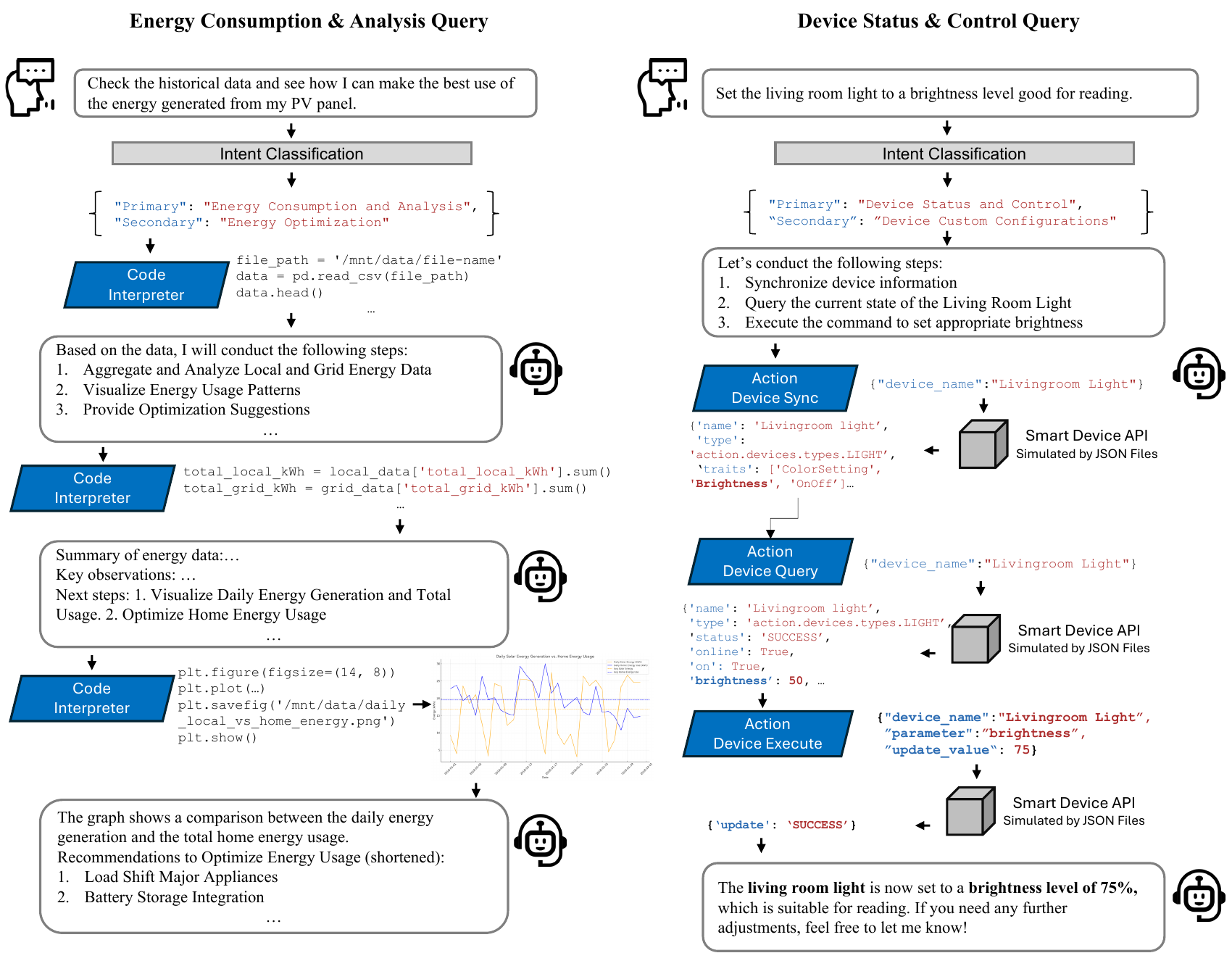}
\caption{Examples of workflow of the AI agent responding to user queries}
\label{fig:workflow}
\end{figure}

One of the major goals of the BEMS AI agent is to provide context-aware energy management insights and suggestions to users. The tested results (Figure \ref{fig:heatmap}) reveal that the AI agent prototype performed well in tasks related to historical energy data analysis (with a response accuracy of 0.9), energy suggestions (0.9), and energy visualization (0.8). However, in tasks related to energy prediction (0.6) and energy optimization (0.7), there is still room for improvement. The result logs for each response to the tested user queries were analyzed to study the workflow of the AI agent in responding to queries. As shown in the example presented in Figure \ref{fig:workflow}, when the user asked how to best utilize the energy generated from the PV panel, the AI agent first conducted intent classification to identify the query's purpose. Following this, the tool "Code Interpreter" was activated to load the stored historical energy data file. Based on the initial data loading, the AI agent outlined the steps required to provide recommendations, then wrote and executed Python code within "Code Interpreter" to generate the initial analysis results. These results enabled the AI agent to identify historical local generation and usage patterns, after which it plotted a figure to visualize these patterns and offered specific recommendations based on the observed trends. This example illustrates how the AI agent follows a logical workflow to systematically provide energy management suggestions and demonstrates its context-awareness by offering recommendations based on actual energy consumption data rather than generic suggestions.

The strong context-aware computing and reasoning abilities of the AI agent are further evident in its handling of energy analysis, optimization, and suggestion tasks. For example, in response to the energy optimization query asking “What are the best practices for optimizing energy use with my home devices?”, the AI agent not only calculated the daily energy consumption for each device but also identified peak usage times for high-consumption devices, such as the dishwasher (with a peak usage at 7:00 p.m.) and the clothes washer (with a peak usage at 9:00 p.m.). The AI agent then recommended shifting the usage of these high-consumption devices to off-peak hours. Another example is the agent's response to the energy suggestion query asking “Can you recommend an energy-saving schedule for my home?” After analyzing the peak usage times of energy-consuming devices and proposing new schedules, the AI agent also automatically activated the tool "action-schedule-CREATE" to adjust the running schedule for the devices. This analysis-suggestion-action process is fully automated, which may appeal to some users for its convenience but may not be favored by others who prefer greater control over their smart devices.

The AI agent did not perform as well in energy prediction tasks, exhibiting long latency (34 seconds), high computational cost (the highest token usage in this category), and the lowest response accuracy (0.6). The tested queries asked about future monthly energy usage for total consumption, specific devices, and energy generation. We compared the ground truth data with the AI agent's predictions to evaluate its accuracy. As we did not specify the methods to be used for prediction tasks, we observed that out of 20 responses, 10 used Random Forest Regression, 6 used Linear Regression, 2 used ARIMA (AutoRegressive Integrated Moving Average) model, and 2 used a simple moving average forecast method. For the machine learning models, the AI agent applied the timestamps as the feature and energy usage data as the target to train the models, all of which were implemented using the "Code Interpreter" tool. Although the models were simple, the accuracy could be improved if specific steps for energy prediction were provided in the instructions. One reason for the low response accuracy is that many responses were evaluated as partially correct (0.5). For example, some responses did not correctly convert energy usage units from kW to kWh but still displayed the results as kWh. Additionally, the prediction models, being too simplistic with limited data (one month of historical energy usage with 15-minute intervals), could not generate accurate predictions, which should be improved with more historical energy data included for analysis.

In addition to energy prediction, the AI agent also failed to provide correct responses for certain test queries due to various issues. In the intent classification step, some energy optimization queries were misclassified as energy suggestion tasks. This is understandable, as there are similarities between the two tasks and their corresponding user queries. In some cases, incorrect parameters were used in the calculations, leading to inaccurate outputs. For instance, for two testbeds in Texas, historical energy consumption data was acquired from January 2018 (a winter month). The data contained records for both 'air' (cooling system) and 'heater' or 'furnace' (heating system). When the user queried about energy consumption related to air conditioning (AC), the AI agent used the incorrect record, resulting in inaccurate results. Another example is the response to the query asking “When are the peak hours of my energy usage over the past month?”, where the expected response should summarize the peak hours from the entire month's data. However, the AI agent oversimplified the task and provided specific peak hours, such as “Here are the peak hours of your energy usage over the past month: 1. January 24th, 21:00 - 22:00, with a total usage of 18.539 kWh. 2. January 12th, 16:00 - 17:00, with a total usage of 17.474 kWh ...”. Although these examples highlight the AI agent's limitations in providing accurate responses in some cases, it is worth noting that with LLM capabilities, the AI agent can understand conversational context. This allows for continuous interaction between users and the AI agent, enabling users to offer further clarification through additional prompts, guiding the AI agent to fully understand the query and generate accurate responses accordingly.

\subsubsection{Cost Management}

An effective means of assisting users in energy management decisions is to compute energy consumption data into direct financial metrics, thereby enhancing clarity around energy-related expenses. To support this objective, the AI agent accesses dynamic energy pricing, accommodating fluctuations in peak and off-peak rates. Moreover, the energy pricing file integrates data on local energy generation, including credits from excess energy returned to the grid, thereby enabling users to maximize both energy efficiency and potential savings. To evaluate the AI agent’s performance in cost management, we examined its responses to user queries related to cost information, predictions, recommendations, and visualizations. Results indicate a process time of approximately 30–40 seconds per query (Figure \ref{fig:heatmap}), with frequent activations of the "Code Interpreter" and "File Search" tools. However, the accuracy of responses was notably lower for cost information (0.3) and prediction (0.5) tasks. Out of the 80 user queries tested, 21 (26\%) responses were accurate, 37 (46\%) were partially correct, and 22 (28\%) were incorrect.

To generate accurate responses for cost-related queries, the AI agent must accurately analyze energy consumption data, retrieve current electricity rates, calculate costs based on consumption and pricing, and synthesize results for user-friendly reporting. Given the multi-step nature of this process, errors in any step can lead to partially or entirely incorrect responses. For instance, in response to the query “Can you provide a breakdown of my past month's energy cost?”, the AI agent incorrectly classified energy generation as a cost rather than a cost-reducing credit, leading to inaccurate reporting. Similarly, when queried about air conditioning (AC) energy consumption, misidentifying the record reference ('air' cooling system instead of 'heater' or 'furnace' heating system) led to erroneous results.

Despite these challenges, the agent demonstrated promising capabilities in scenarios where it provided correct responses. For example, in response to the query “What device incurred the highest charge on my energy bill last month?”, the AI agent accurately calculated costs for all home appliances and identified the electric vehicle charger as the device with the highest charge. In another case, for energy cost prediction the query “What is the estimated cost of my energy usage for next month?”, the AI agent employed random forest regression to forecast energy usage and then calculated an estimated cost of \$51.73, closely aligning with the actual cost of \$52.86. These correct responses highlight the AI agent's potential in cost management, if its processes can be standardized and rigorously validated for accuracy. Enhancements in instructional design and system architecture are necessary to further improve the AI agent's accuracy of cost management.

\subsubsection{Data Visualization}

\begin{figure}[ht]
\centering
\includegraphics[width = 1.0 \linewidth]{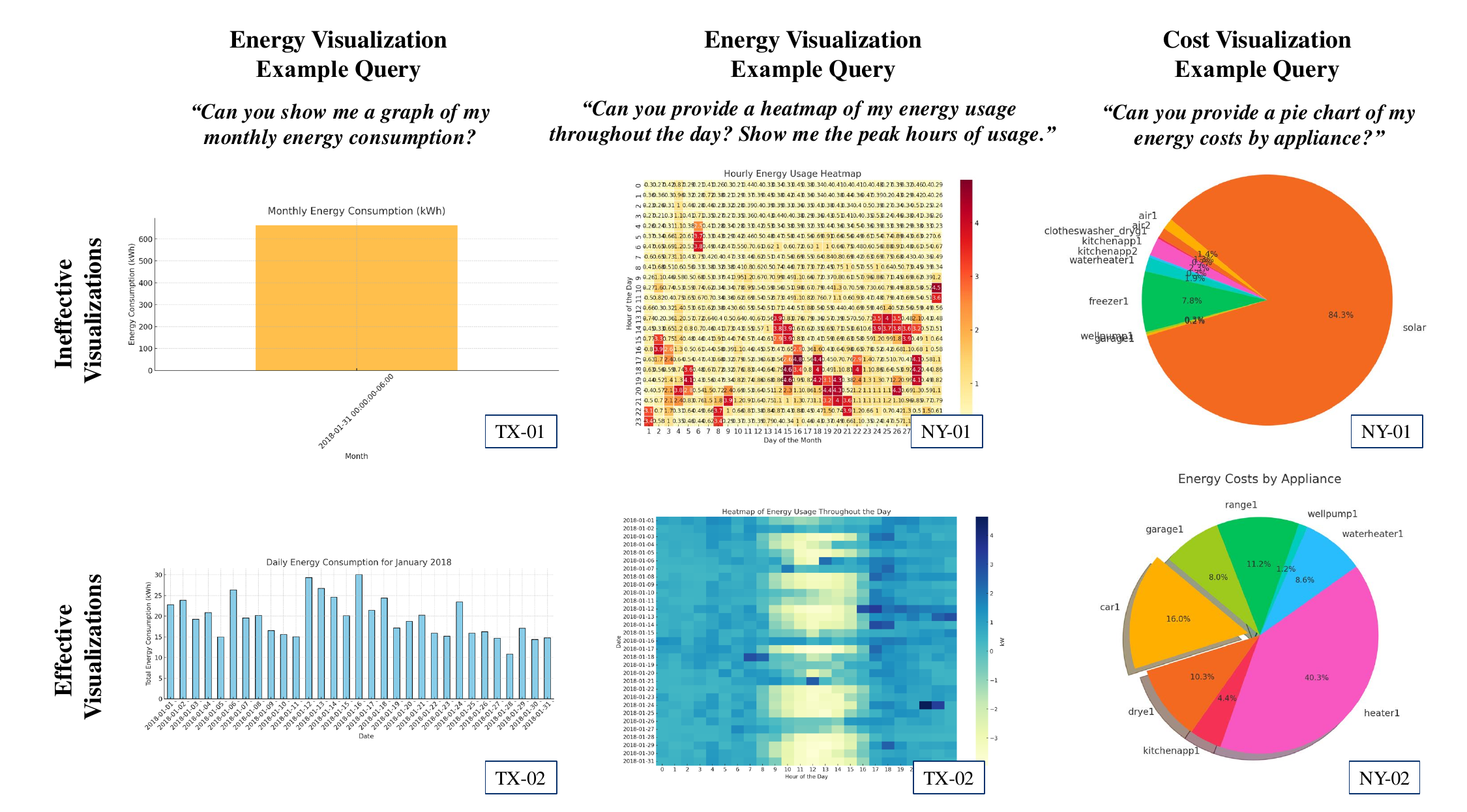}
\caption{Examples of efficient and inefficient visualization plots generated by AI agents}
\label{fig:vis_compare}
\end{figure}

Using the "Code Interpreter" tool, the AI agent executes Python code to generate data visualizations in response to user queries. To evaluate this capability, we designed queries focused on energy visualization and cost visualization to assess its support for users' decision-making processes. Of the 40 visualizations produced in response to these queries, 3 were entirely incorrect, 21 were partially correct, and 16 were accurate. We categorized the partially correct visualizations as inefficient, while accurate visualizations were classified as efficient, as demonstrated in the examples shown in Figure \ref{fig:vis_compare}.

The inefficient visualizations provided insufficient information for effective energy management. For instance, in response to the query “Can you show me a graph of my monthly energy consumption?”, the response from the TX-01 testbed presented a single bar chart representing total monthly energy consumption. This visualization failed to offer meaningful insights for users. In contrast, the response from the TX-02 testbed generated a series of daily energy consumption bars, illustrating fluctuations and peaks across the month, and offering more actionable information for users. Similarly, in response to the query “Can you provide a pie chart of my energy costs by appliance?”, the response from the NY-01 testbed included energy credits within the energy cost breakdown, without direct user question. Conversely, the visualization generated for the NY-02 testbed provided a detailed and accurate pie chart, meeting user needs more effectively. Improving the generation of efficient visualizations will require refining the AI agent’s instructions and workflow design. Specifically, when user queries are general or vague, the AI agent should seek further clarification (e.g., plot type, data range, etc.) to produce more informative visualizations conducive to decision-making.

In addition to queries that explicitly request visualizations, the AI agent autonomously generated visualizations for certain unspecified queries, illustrating its proactive decision-support capabilities. In 33 instances, the AI agent created visualizations without explicit prompts, including 8 instances related to cost management and 25 related to energy consumption analysis. For example, for the query “Can you recommend an energy-saving schedule for my home?”, for the TX-01 testbed, it generated a heatmap displaying average energy consumption across weekday hours. This visualization effectively highlighted peak usage periods, enhancing the agent’s explanation of the recommended energy-saving schedule. Although some of the unanticipated visualizations are efficient visualizations that can contribute to AI agent's responses, some of the visualizations are inefficient and less informative. The mechanics of generating data visualization without specification need to be further standardized. In addition, the impact of unanticipated visualizations on user satisfaction with the AI agent's responses needs further investigation in future studies.

\subsubsection{Device Control and Scheduling}

Current home energy management systems generally lack direct control capabilities for smart appliances. To address this gap, the BEMS AI agent includes functionality to control smart devices. To evaluate the AI agent’s device control and scheduling capabilities, we tested various queries involving device control and scheduling tasks (as shown in Table \ref{tab:intent_categories}). When the AI agent receives a query related to smart devices, it processes the query through the workflow illustrated in Figure \ref{fig:flow}.In the prototype, following intent classification, which identifies the target device and the necessary commands, the AI agent follows a sync-query-execute workflow to stepwise control the devices. It first initiates the "action-devices-SYNC" tool to gather device information and determine the relevant control traits. It then calls the "action-devices-QUERY" tool to ascertain the device’s current status, and subsequently uses the "action-devices-EXECUTE" tool to send control commands. Similarly, for device scheduling, the agent employs tools such as "action-schedules-SYNC", "action-schedules-CREATE", and "action-schedules-CHANGE" to establish automation schedules. All tools are connected with a simulated smart device API to simulate the device control process. This structured workflow ensures that the AI agent’s actions module operates with contextual awareness, minimizing hallucinations that might occur in LLM applications.

The evaluation of the AI agent’s responses to the tested queries is summarized in Figure \ref{fig:heatmap}. The results indicate high accuracy in device status and management responses, with average response accuracy scores above 0.8. For single-device status checks and general operations, process time was efficient, averaging below 14 seconds. However, group device management and custom configurations took longer, with average processing times of 25 seconds and 32 seconds, respectively, due to the need for multiple tool calls (typically four per group operation). In terms of prompt token requirements, single device checks and operations required fewer tokens compared to group management and customization. For scheduling tasks, the AI agent demonstrated high accuracy in general scheduling and conditional automation (above 0.8). However, scheduling information management exhibited a lower average accuracy score (0.6), which can be attributed to a limitation in the prototype implementation rather than limitations in the AI agent’s reasoning or tool-selection capabilities. Specifically, the simulated API used in the evaluation did not consistently return complete scheduling information across devices due to an implementation inconsistency, leading to incomplete inputs for the agent and, consequently, reduced response accuracy in these cases.

The sync-query-execute workflow, designed to identify available devices, control-specific traits, and the required commands and parameters, considerably enhances the AI agent’s contextual awareness. For example, in response to the query "Set the living room light to a brightness level good for reading," the AI agent identified the brightness adjustment option for the living room light, detected the current brightness level (50), and adjusted it to 75 to enhance readability (Figure \ref{fig:flow}). In another example, for the scheduling query "Set my car charger to be on during off-peak hours," the AI agent used the "File Search" tool to identify off-peak hours (00:00 - 17:00, 20:00-24:00) from an energy pricing document, then employed the "action-schedule-CREATE" tool to configure the EV charger schedule accordingly. Notably, in these examples, the AI agent inferred parameters (e.g., brightness, peak hours) without explicit user input, effectively tailoring responses to user needs.

Enhanced contextual awareness also improves control accuracy of the smart devices. For instance, in our simulated smart devices API, the kitchen kettle’s “online” status was set to “false,” indicating it was offline. For the query "Turn on my kitchen kettle, and set the temperature to 80 degrees," the AI agent correctly identified the offline status and suggested that the user needs to check the system and network settings. Similarly, in response to the device scheduling query "Turn off the TV when no one is in the room," the agent detected the absence of a TV in the appliances list, thereby preventing an erroneous action. In the TX-01 testbed, after identifying the unavailability of a TV, the AI agent adapted by scheduling adjustments for other devices (lighting, AC) in the living room when no occupants were detected, demonstrating notable adaptability. However, for ambiguous queries, such as the scheduling query "Set the AC to 21 degrees when I go to sleep," the AI agent sought additional clarification to specify the actuation time, indicating challenges in interpreting vague requests. Striking a balance between minimizing user prompts and maximizing clarity and accuracy in smart device control remains an area for further investigation.

\subsubsection{Memory}

The short-term memory, which primarily involves conversational history between the user and the AI agent, can be supported by the capabilities of LLM itself. This enables users to provide follow-up prompts and refine requests if the initial response does not meet their expectations. Our study specifically assesses the long-term memory capabilities of the AI agent. We simulated the AI agent’s long-term memory in a JSON file format, including user preference information such as "The user prefers the AC set to 21 degrees at bedtime" and "The user prefers the bedroom lights to be turned on at sunset". To evaluate this functionality, we tested specific tools designed for memory operations, including "action-memory-SYNC", "action-memory-CREATE", and "action-memory-CHANGE". The evaluations (Figure \ref{fig:heatmap}) demonstrated a quick process time (averaging 9 seconds for memory information retrieval and creation, and 17 seconds for memory management), along with high response accuracy (average scores above 0.9) for memory-related tasks. However, the low intent classification execution rate of memory-related tasks (with an average score below 0.3 as shown in Figure \ref{fig:heatmap}) indicates that for memory-specific queries, the AI agent often bypasses intent classification and directly calls memory-related tools, simplifying the process.

Beyond queries explicitly related to memory, we observed that memory tools were used in 15 instances where memory was not specified in user queries. These cases included 10 responses related to device control and scheduling, and 5 related to energy and cost management. For example, in response to the query "Based on historical data, what are some habits I can adopt to reduce energy usage?", the AI agent synchronized its memory of the user’s existing device control habits before generating tailored suggestions. For device scheduling tasks, the AI agent can also generate and store specific device scheduling information as long-term memory. Although this memory creation feature was not frequently observed, it shows promise for enhancing the AI agent’s effectiveness in relevant contexts. Further research is needed to determine which memory entries are beneficial and should be retained, and which might be redundant and warrant deletion to optimize the AI agent’s interaction with users.

\subsubsection{General Information \& Support}

Test queries in this category generally do not require the AI agent to use specific tools to respond to users. Accordingly, as shown in Figure \ref{fig:heatmap}, the average tool call count, defined as the mean number of function or tool calls per query, is low for the subcategories of System guidance and tutorials and FAQs and general queries, with values below 0.3. This indicates that most queries in these subcategories were answered without invoking any tools. In contrast, Troubleshooting and support queries typically require interaction with device-related tools to diagnose system states and propose corrective actions, resulting in a higher average tool call count of 2.2 per query. The response accuracy for queries in this category averaged above 0.9, indicating high accuracy and reliability in addressing general information queries to support users. For instance, in response to the system guidance and tutorial query "Can you guide me through the process of creating a custom schedule for smart devices?", the AI agent provided a detailed, step-by-step tutorial for creating schedules and used relevant tools for smart appliance scheduling. In response to the troubleshooting and technical support query "My kettle doesn't work, can you help me check it?", the AI agent used the "action-devices-QUERY" tool to retrieve the current device status. In this case, the agent identified that the device’s “online” status was “off” and suggested that the user check the network connection of the kitchen kettle. In the case of FAQs and general queries, the query "Can you explain the different modes available in my AC?", the AI agent reviewed the modes available for the air conditioning unit, identified options such as off/on/cool/heat/fan-only/eco, and explained each mode to the user.

\section{Discussion}
\label{sec:discussion}

\subsection{Correlation Among Evaluation Metrics}

\begin{figure}[t]
\centering
\includegraphics[width = 0.9 \linewidth]{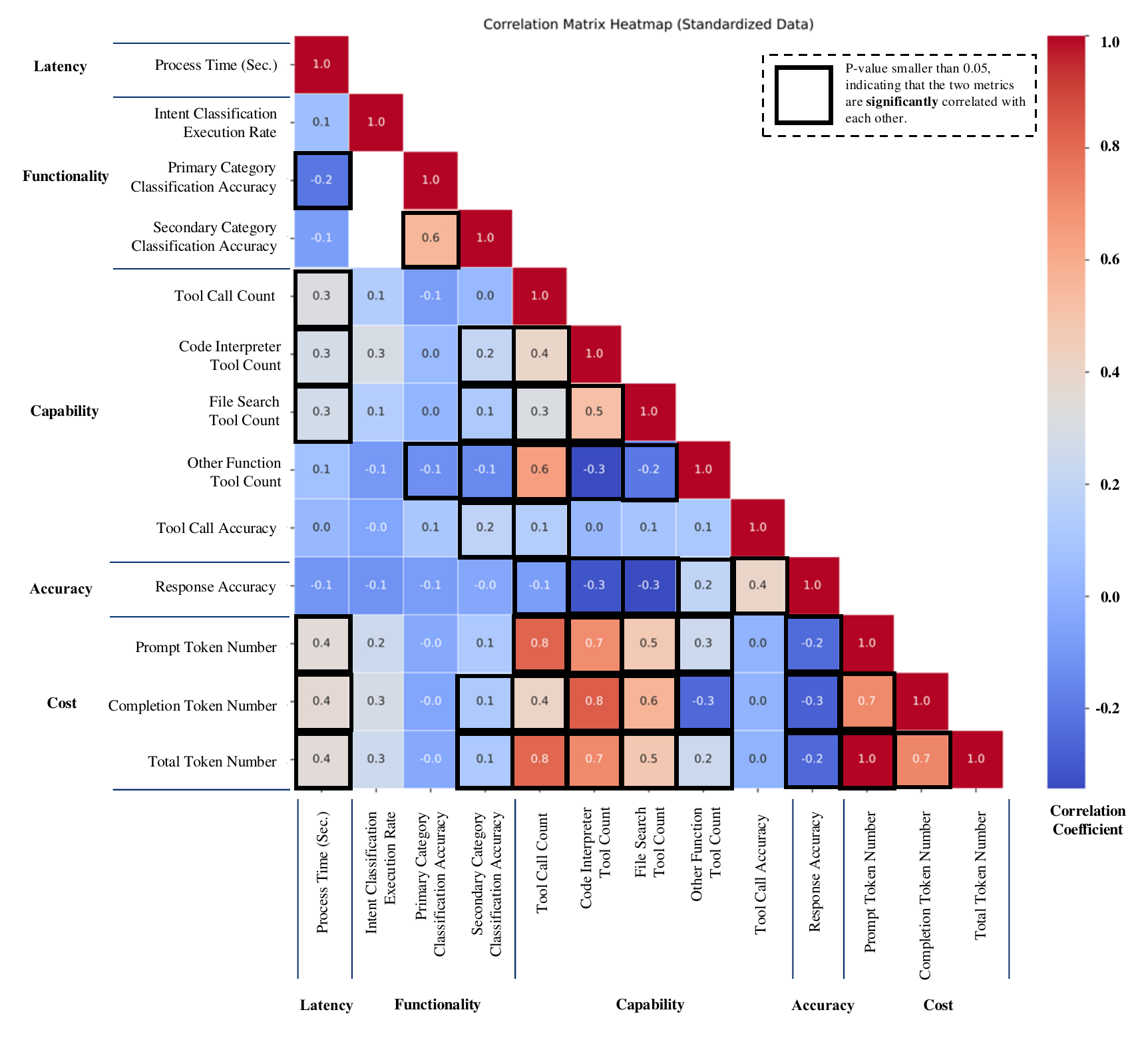}
\caption{Heatmap built based on correlation analysis of the evaluation metrics}
\label{fig:corre}
\end{figure}

Following the evaluation of the AI agent prototypes, we conducted a correlation analysis across various performance metrics to better understand interdependencies and the potential trade-offs in the responses. Figure \ref{fig:corre} presents a heatmap illustrating the correlation coefficients between these metrics. Certain cells in the heatmap are empty due to the absence of intent classification in some user query responses. As expected, the analysis reveals a positive correlation between processing time and several metrics related to computational load, such as the number of tool calls, activation of specific tools (e.g., code interpreter and file search), and prompt token count. In practical terms, more resource-intensive tasks such as detailed energy analysis or cost computation involve higher computational overhead. In addition, functional metrics such as intent classification rates and accuracy exhibit limited correlation with other evaluation metrics, except for a positive correlation between the secondary intent classification accuracy and both tool call frequency and token usage. This suggests that while intent classification is generally consistent across task types, secondary classifications relate to more detailed prompts or tool activation, particularly for nuanced or multi-step queries where additional context is required to respond to users.

For metrics associated with the analytical capabilities, the tool call frequency (especially for the code interpreter and file search tools) aligns closely with prompt, completion, and total token usage. This pattern particularly occurred in energy and cost management tasks, where the AI agent often leverages these tools extensively to parse and analyze data. Furthermore, response accuracy demonstrates a significant negative correlation with both tool call frequency and token usage potentially attributed to the increased complexity of tasks, which often introduces uncertainties or requires extensive data interpretation, resulting in relatively lower accuracy scores. Conversely, simpler tasks---such as device control, scheduling, and memory management---demand fewer computational resources and exhibit higher response accuracy. These findings highlight the importance of optimizing resource allocation and tool activation strategies for complex tasks.

\subsection{Comprehensive and multidimensional BEMS AI Agents assessment}

This study presents a comprehensive assessment of a proposed large language model (LLM)-based BEMS AI agent, designed to provide context-aware assistance using energy consumption data and the operational characteristics of building systems and smart appliances. Previous research has examined the use of LLMs in data-driven tasks within building energy management \cite{zhang2024generative} and in optimizing power dispatch \cite{zhou2024elecbench}. Additionally, LLMs have been applied in smart building environments to enhance user interaction, as demonstrated by systems such as SAGE (Smart Home Agent with Grounded Execution) \cite{rivkin2023sage} and Sasha (Smarter Smart Home Assistant) \cite{king2024sasha}, which address general or ambiguous user queries for device control. Distinct from these prior efforts, this study leveraged a comprehensive set of queries that integrates both analytical energy management functions, such as consumption analysis, usage prediction, and cost estimation, and the actionable control and scheduling of smart appliances within an interactive system.

Our assessments include the evaluation of the BEMS AI agent’s performance in practical scenarios where users engage with both informational and control-oriented functionalities. Specifically, this study assesses the agent’s capacity to provide actionable insights via energy data analysis, predict future energy demands based on historical patterns, estimate costs across operational scenarios, and provide device control commands. Testing on four real-world residential energy datasets validates the framework's robustness, providing a reference for future research. While the prototype exhibited promising outcomes in terms of latency, cost, and accuracy, there remains room for improvement, particularly regarding latency. Compared with previous studies, where smart speakers required an average of 7.43 seconds for information queries and 8.27 seconds for IoT control tasks \cite{mun2020smart}, and LLMs averaged 15 seconds for complex reasoning tasks \cite{chen2024livemind}, our prototype takes longer to perform complex tasks. Specifically, energy analysis tasks average 27 seconds, cost management tasks take 34 seconds, device control requires 19 seconds, and device scheduling and automation are completed in 14 seconds (Table \ref{tab:metrics_time_cost_acc}). This latency suggests the need for improvements to facilitate smoother user interactions. Assessments revealed 86\% accuracy in responding to device status and control queries, 74\% in device scheduling and automation tasks, and 77\% in energy consumption and analysis. However, accuracy in cost quantification tasks was lower at 49\%, indicating that further development is needed for complex computational and reasoning tasks. In terms of task completion cost, task complexity is an influential factor as expected. For instance, energy consumption \& analysis queries incur a cost of approximately \$0.095 per response, while cost management tasks cost around \$0.141 per response. Simpler tasks, such as device status \& control, average \$0.075, and device scheduling \& automation cost about \$0.056 per response. These estimates reflect pricing at the time of experimentation and may vary substantially with model selection and future pricing changes. Although LLM model costs may decrease as models advance, these evaluations offer a baseline reference for building energy management task expenditures---an aspect not extensively explored in previous studies.

This study contributes to the field from different perspectives. First, it moves toward establishing a standard way for evaluating LLM-based AI agents in smart building applications. Second, it provides a structural foundation for designing AI agents with enhanced contextual understanding, enabling robust responses to complex queries requiring both analytical and operational outputs. Finally, the quantified assessments serve as a baseline for future research, facilitating comparative analyses of BEMS AI agent performance across different architectural configurations, energy scenarios, and user interaction models.

\subsection{Limitations and Future Studies}

\textbf{User Interactions:} In this study, our evaluations provide a baseline assessment of the framework's current performance. However, to enable broader integration and practical application in everyday scenarios, it is essential for the AI agent to maintain robustness across a wider range of user queries \cite{chang2024survey}. Further research is therefore needed to explore this dimension and investigate user-agent communication in extended, multi-round interactions. These areas call for end-user studies to improve understanding of natural interaction patterns, which could enhance interaction quality, making it more intuitive and better suited for daily smart applications. User-centered evaluation of AI agent output, including user satisfaction, engagement, usability, likeability, and acceptability is another important direction of research \cite{giudici2024designing, chang2024survey}. Lastly, proactive interactions initiated by the BEMS AI agents represent another promising area for future work. Prior studies indicate that proactive suggestions from AI assistants can help users in efficient energy use \cite{he2022ai}. To this end, BEMS AI agents should be capable of identifying energy consumption patterns and offering timely suggestions that support users' goals.

\textbf{Safety and Privacy:} Safety and privacy are critical considerations in the deployment of AI-driven systems within built environments. These systems are inherently vulnerable to breaches and errors. Given that AI agents may directly control smart devices through an Action module, it is essential to ensure that users retain ultimate control over these systems to ensure desired outcomes including those that could affect user experiences and well-being. The challenge of ensuring that language models consistently deliver high-quality responses has been noted in prior research \cite{chang2024survey}. Privacy concerns are also important, particularly due to the sensitivity of user data. Equipped with meters and sensors within smart environments, an AI agent can analyze building energy usage patterns which could be correlated to user activities \cite{juttner2024chatanalysis}. Addressing these dimensions is an important research direction in developing smart building AI agents \cite{ramokapane2022privacy}.

\textbf{Agentic Workflow with Multi-Agents:} The use of agentic workflows and multi-agent systems in LLM-supported environments presents an opportunity to enhance accuracy in smart building applications. Current tools leverage code generation to support the analyses, which may lack the stability required for reliable, repeatable outcomes. This limitation was also observed in a study that applied GPT-4 for energy load prediction and anomaly detection, which highlighted the challenges in maintaining consistent model performance \cite{zhang2024generative}. Developing an agentic workflow or multi-agent system may offer a solution by distributing specialized tasks across multiple agents, each tailored to handle specific processes \cite{wu2023autogen}. This approach may improve task allocation and enhance scalability and robustness. Additionally, further research is warranted into automating the creation of AI agent profiles and user profiles, reducing the need for manual intervention.

\textbf{Infrastructure-level impact of AI Agents:} Understanding the infrastructure-level impact of AI agents, particularly those powered by LLMs, constitutes an important research area \cite{ludvigsen2023carbon}. Some studies have attempted to estimate the impact of AI models operations including the significant GPU energy use \cite{luccioni2024power}. Water consumption is another critical factor, particularly for data centers \cite{luccioni2023estimating}. Therefore, it would be counterproductive for BEMS AI agents, designed to assist users in improving energy efficiency, to contribute to the total energy use in their operation. Therefore, life cycle evaluation of the BEMS AI agents is important to ensure that overall they achieve energy efficiency goals. 

\section{Conclusion}
\label{sec:conclusion}

This study presents a framework and benchmark assessment for Large Language Model (LLM)-based Building Energy Management System (BEMS) AI agents that could support users in managing energy use and smart devices through context-aware, intuitive, and natural language interactions. The proposed BEMS AI agent framework consists of three primary modules---perception, brain (central control), and action, facilitating the collection, processing, and management of energy data alongside the operation of smart appliances. Leveraging LLMs for enhanced data-driven contextual awareness, the agent can perform tasks such as energy consumption analysis, cost estimation, and device automation by using contextual operational data.

Building on the conceptual framework, a prototype was developed and evaluated through extensive simulated testing using a benchmark set of 120 user queries spanning multiple primary and secondary task categories. The evaluation leveraged data from four real-world residential datasets and adopted a multidimensional set of metrics, including latency, functionality, capability, accuracy, and computational cost. The results demonstrated the strong potential of the proposed BEMS AI agent to deliver effective, contextually relevant, and generalizable energy management support. A series of ANOVA tests conducted across different residential settings showed no statistically significant differences in performance metrics (p-values larger than 0.05), indicating consistent behavior and supporting the generalizability of the proposed framework across different building configurations.

A closer examination of individual metrics showed that the AI agent achieved high response accuracy for several task categories, including device status and control (86\%), memory-related tasks (97\%), and general information and support queries (98\%). Performance remained promising for energy consumption analysis (77\%) and device scheduling and automation (74\%), while more complex tasks such as cost management exhibited lower accuracy (49\%), reflecting the challenges associated with multi-step data interpretation and analytical reasoning. Latency analysis showed an average process time of 23 seconds per query, highlighting the need for improvement in real-time deployment scenarios through streamlined workflows and more efficient models. In terms of functional performance, the prototype achieved classification accuracies of 91\% for primary task categories and 76\% for secondary categories, alongside a tool call accuracy of 94\%. From a computational perspective, the agent required an average of 29,467 tokens per query. More resource-intensive tasks, such as energy analysis and cost management, required higher token usage (35,625 and 52,761 tokens, respectively), whereas device control and automation tasks were comparatively less demanding. Correlation analysis across evaluation metrics highlighted trade-offs between task complexity, response latency, and computational cost, emphasizing the importance of optimizing tool utilization and reasoning workflows.

This study presents a benchmarking assessment that encompasses both analytical and operational functionalities, moving toward establishing a standard for evaluating LLM-based AI agents in smart building energy management systems. Future work could aim to enhance user interactions in real-world settings, address user perspectives, and explore agentic workflows using multi-agent approaches.


\section{Generative AI Use}

During the preparation of this work, the authors employed generative AI as an assistive tool to refine and polish the text. Subsequent to utilizing this tool/service, the authors reviewed and edited the content, ensuring its accuracy and relevance, and take full responsibility for the content of the publication.

\appendix

 \bibliographystyle{elsarticle-num} 
 \bibliography{cas-refs}





\end{document}